\documentclass[journal=eds]{CUP-JNL-DTM}%

\usepackage{graphicx}
\usepackage{multicol,multirow}
\usepackage{amsmath,amssymb,amsfonts}
\usepackage{mathrsfs}
\usepackage{amsthm}
\usepackage{rotating}
\usepackage{appendix}
\usepackage{ifpdf}
\usepackage[T1]{fontenc}
\usepackage{newtxtext}
\usepackage{newtxmath}
\usepackage{textcomp}
\usepackage{xcolor}
\usepackage{lipsum}
\usepackage[colorlinks,allcolors=blue]{hyperref}
\usepackage{subcaption}
\usepackage{comment}
\usepackage{xr}
\theoremstyle{definition}

\numberwithin{equation}{section}

\jname{Environmental Data Science}
\articletype{Methods Paper}
\jyear{2026}

\begin{document}

\begin{Frontmatter}

\title[Article Title]{Prototype-based Explainable Neural Networks with Channel-specific Reasoning for Geospatial Learning Tasks}

\author[1,2]{Anushka Narayanan}
\author[1,2,3]{Karianne J.\ Bergen}

\authormark{Narayanan and Bergen}

\address[1]{\orgdiv{Department of Earth, Environmental and Planetary Sciences}, \orgname{Brown University}, \orgaddress{\city{Providence}, \postcode{02906}, \state{RI},  \country{USA}}}

\address[2]{\orgdiv{Data Science Institute}, \orgname{Brown University}, \orgaddress{\city{Providence}, \postcode{02906}, \state{RI},  \country{USA}}}

\address[3]{\orgdiv{Department of Computer Science}, \orgname{Brown University}, \orgaddress{\city{Providence}, \postcode{02906}, \state{RI},  \country{USA}}}

\keywords{neural networks, explainable artificial intelligence, machine learning, geospatial data analysis, raster data analysis} 


\abstract{Explainable AI (XAI) is essential for understanding machine learning (ML) decision-making and ensuring model trustworthiness in scientific applications. Prototype-based XAI methods offer an intrinsically interpretable alternative to post-hoc approaches which often yield inconsistent explanations. Prototype-based XAI methods make predictions based on the similarity between inputs and learned prototypes that represent typical characteristics of target classes. However, existing prototype-based models are primarily designed for standard RGB image data and are not optimized for the distinct, variable-specific channels commonly found in geoscientific image and raster datasets. In this study, we develop a prototype-based XAI approach tailored for multi-channel geospatial data, where each channel represents a distinct physical environmental variable or spectral channel. Our approach enables the model to identify separate, channel-specific prototypical characteristics sourced from multiple distinct training examples that inform how these features individually and in combination influence model prediction while achieving comparable performance to standard neural networks. We demonstrate this method through two geoscientific case studies: (1) classification of Madden Julian Oscillation phases using multi-variable climate data and (2) land-use classification from multispectral satellite imagery. This approach produces both local (instance-level) and global (model-level) explanations for providing insights into feature-relevance across channels. By explicitly incorporating channel-prototypes into the prediction process, we discuss how this approach enhances the transparency and trustworthiness of ML models for geoscientific learning tasks.}

\end{Frontmatter}

\section*{Impact Statement}
This work introduces an explainable AI (XAI) approach where machine learning models use intrinsically transparent reasoning mechanisms for tasks with complex geoscientific data. Our method explains model predictions by comparing inputs to prototypes which are typical patterns representing key environmental features identified by the model. Designed for multi-channel data where each channel represents distinct environmental features like climate variables or satellite bands, this approach identifies how each channel contributes to predictions for individual instances as well as across the entire model. We apply this in two geoscientific case studies, a climate pattern classification task and a land-use mapping task. This approach reveals how different variable-specific information influences model decisions, improving transparency of deep learning in the geosciences.

\section{Introduction}

Deep learning (DL) has been an increasingly popular tool for modeling complex phenomena in Earth and climate science \citep{rolnick2022tackling, mcgovern2023review, wang2023scientific} due to its faster computational speeds, improved efficiency, and robust performance when handling large geoscientific datasets \citep{schultz2021can}. Deep neural networks typically learn a latent representation of input data (\textit{embeddings}); however, this process is often a black box offering little insight into ``how'' or ``why''  model predictions are made \citep{de2023machine, molina2023review}, which is often essential for geoscientists \citep{mamalakis2020explainable}. The complex nature of deep neural networks, characterized by multiple nonlinear layers and a large set of parameters, makes visualizing and interpreting its internal representations difficult \citep{zhang2018visual, molnar2022}. Explainable AI (XAI) addressed this challenge by providing a set of tools and techniques that interrogate the model decision-making process to provide insights into how or why a model makes specific predictions \citep{nauta2023anecdotal}. XAI tools can increase model transparency and help verify whether model predictions were made in line with scientific reasoning \citep{mcgovern2019making}. 

In the geosciences, XAI is most commonly implemented through a category of \textit{post hoc} techniques that are applied to trained black box models after inference \citep{yang2024interpretable}. Attribution-based post hoc approaches such as SHapley Additive exPlanations (SHAP) \citep{lundberg2017unified} and Layer-Wise Relevance Propagation (LRP) \citep{bach2015pixel} are frequently used to assess the influence of input features on model predictions \citep{van2025iseflow, griffin2022predicting, toms2020physically}. However, growing evidence indicates these techniques can produce inconsistent or unreliable explanations, with results varying across methods \citep{neely2021order, zhou2021evaluating}. For example, \citet{mamalakis2022investigating} identify limitations in a climate prediction context, including challenges in distinguishing positive from negative feature contributions. Likewise, \citet{bommer2024finding} report that different post hoc methods can generate conflicting explanations for the same prediction instance in a separate climate application.

\begin{figure}[htb!]%
\FIG{\includegraphics[width=0.9\textwidth]{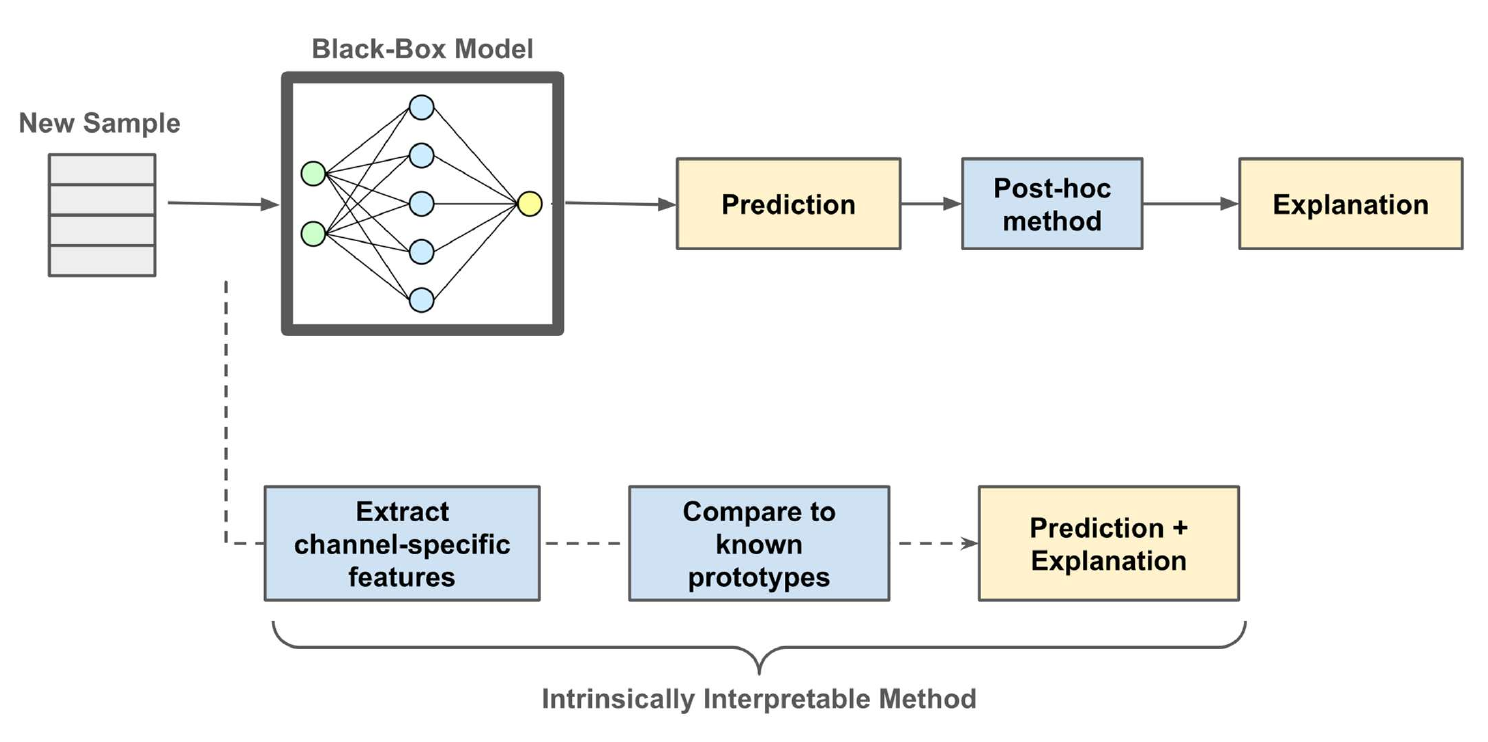}}
{\caption{Diagrammatic representation of (upper): a standard neural network with post hoc XAI applied to a trained model to produce an explanation and (lower): an intrinsically interpretable alternative model with prototype-based reasoning to produce an explanation}
\label{outline}}
\end{figure}

An alternative approach is \textit{intrinsically interpretable} XAI -- methods that modify the neural network architecture to directly incorporate interpretable reasoning \citep{rudin2019stop} and bypass the need to use post-hoc XAI tools altogether \citep{balci2023intrinsically, adey2024exploration, o2025moving}. An example is a class of methods referred to as \textit{prototype networks} \citep{li2018deep}. First introduced for computer vision tasks, prototype networks make predictions by comparing the similarity between inputs and a set of learned ``prototypes'' consisting of patterns that are representative of each class of the training data \citep{chen2019looks}. Unlike standard neural networks, which directly pass embeddings into a classifier, prototype networks pass a vector representing the similarity of the input to class-specific prototypes into the classifier to make a final prediction \citep{davoudi2021toward}. An outline of the model methodology is shown in Figure \ref{outline}. Prototype-based XAI methods are relatively underutilized in climate and Earth science and present opportunities to provide novel and relevant model explanations to the user from the model's built-in reasoning process \citep{barnes2022looks, narayanan2024prototype}. 
A key obstacle in applying prototype-based XAI tools to scientific datasets is that they are designed for and tested on natural image classification benchmark tasks. Geoscientific datasets exhibit distinct characteristics from natural images: multi-scale spatiotemporal patterns, ambiguous object boundaries, non-RGB image channels, highly correlated features \citep{krell2025influence} and multi-spectral data \citep{rolf2024mission}, which may require adaptations to produce scientifically useful and relevant explanations \citep{narayanan2024prototype}. For example, absolute location is often important for geospatial data, unlike for natural images. \citet{barnes2022looks} addressed this by introducing \textit{location scaling} to incorporate the notion of the importance of the location of climate features into a prototype network architecture for a Madden Julian Oscillation (MJO) phase classification task.

In this work, we focus on another key difference between geoscientific and natural image data: the relationship and meaning of individual channels. Unlike natural images, where three RGB channels jointly encode a single feature, the color of each pixel, geospatial raster data consists of separate channels containing gridded geophysical variables such as elevation, temperature, meridional wind or spectral channels beyond RGB. We refer to these datasets collectively as geoscientific ``image" data due to their pixel-based structure. Each channel in a geoscientific image may contain task-relevant information \citep{hilburn2020development} unlike natural images, where relevant information emerges when RGB channels are combined. Prior work by \citet{ghosal2021multi} on prototype networks for multi-variable times series demonstrated that pertinent information elicited from each individual time series channel could enhance explanations of model decisions. We hypothesize that methods which capture relevant spatial patterns within each individual geospatial channel can provide greater insights into model reasoning. To our knowledge, no studies have explored the development of channel-specific prototype based methods for geoscientific use-cases. 

In this paper, we introduce a novel prototype-based XAI method that generates predictions and associated channel-based explanations that account for key channel-specific patterns in the input data. Building on \citet{chen2019looks}'s prototype network intended for RGB image use and \citet{barnes2022looks}'s location-scaled prototype network, we extend prototype networks to capture channel-specific prototypes for multi-channel geoscientific image-based use-cases and generate channel- and location-specific insights. Our method can produce both \textit{local} explanations \citep{saranya2023systematic} that provide the model's reasoning for the prediction of a specific instance and \textit{global} explanations that describe the importance of the channel-specific patterns to provide an understanding of the overall model reasoning, which is relatively under-discussed in prototype-based XAI literature \citep{nauta2021neural}.

We present three case studies that demonstrate the versatility of the method with multi-channel geoscientfic datasets by highlighting channel-specific information relevant to each individual task. The first case study features a classification task using a synthetically generated multi-channel dataset based on the MNIST image classification benchmark. The second experiment presents a Madden Julian Oscillation (MJO) climate prediction task using multi-channel climate datasets consisting of gridded climate variables, following the case study featured in \citet{barnes2022looks}. Third, we demonstrate this method on a multispectral remote sensing classification task, an application which, to our knowledge, has not yet been explored using prototype-based XAI. We also examine whether prototype-based explanations can be derived from pre-trained neural networks whose encoders were originally trained as standard black-box models without prototypes.

Overall, we discuss how we establish a channel-specific prototype framework that offers novel local and global explanations and provides new opportunities to improve interpretability and trustworthiness in geoscientific DL applications.

\section{Method}

In this section, we describe the overall structure of the model architecture and detail the general training procedure and hyperparameter tuning necessary to implement the model. The model architecture (Figure \ref{fig:diagram}) consists of the feature embedding layers, referred to as the encoder ($E$), the prototype learning layer ($P$), and final classification layer ($L$). The encoder allows the model to extract relevant features from the input data, similar to any standard neural network. The prototype learning layer identifies prototypical patterns for each class within the embedding space. Similarity values are computed between the test sample's embedding and the learnt prototypes. These similarity values are input into the final classification layer to output a prediction. By compelling the model to compare the test sample to a set of known, interpretable prototypical patterns, the model is able to simultaneously produce a prediction along with an similarity-based reasoning thus generating \textit{local} explanations for a specific test sample. We also discuss how the method can be used to generate a \textit{global} explanation of the reasoning mechanisms and relevant features that contribute to the model's overall understanding of the dataset.

\subsection{Model Architecture}
\begin{figure}[htb!]
    \centering
    \includegraphics[width=0.83\linewidth]{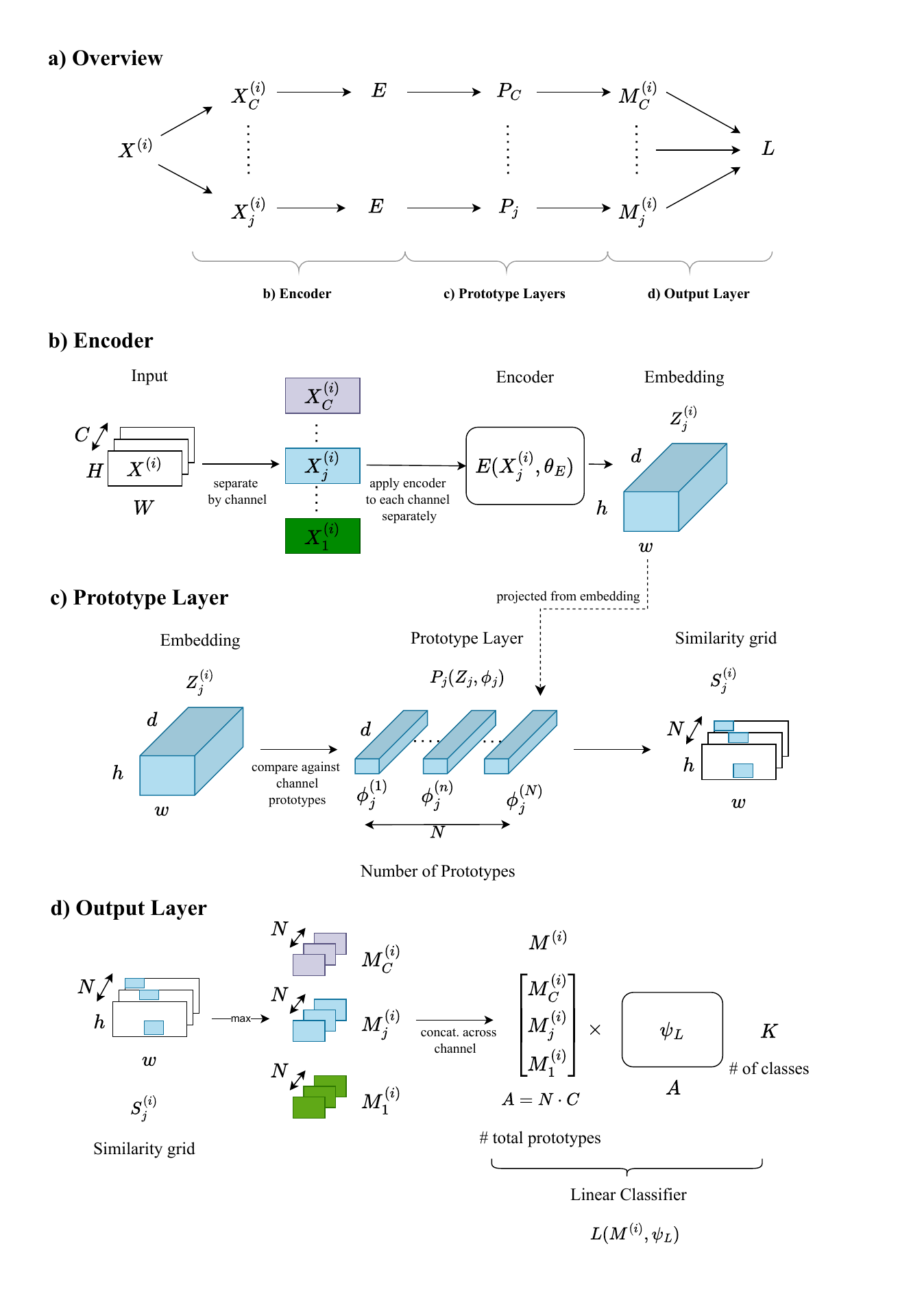}
    \caption{(a) High-level architecture overview of the channel-specific prototype method. (b) Encoder: Input sample, $X^{(i)}$, is separated channel-wise and processed through the encoder, $E$, to produce an embedding for each individual channel (a single channel's embedding is displayed here for simplicity). (c) Prototype Learning Layer: Each channel-specific embedding, $Z_j^{(i)}$, is processed through its respective prototype layer, $P_j$, and compared against the set of $N$ prototypes for this channel to produce $N$ similarity grids, $S_j^{(i)}$. (a single channel's operations are displayed here for simplicity) (d) Output Layer: Maximum similarity values for each prototype are obtained and concatenated across all channels resulting in a $N \cdot C$ length vector and fed through the linear classifier to produce a prediction.}
    \label{fig:diagram}
\end{figure}

Our proposed method follows a similar prototype-based neural network architecture developed by \citet{chen2019looks} with location scaling \citep{barnes2022looks} that separates the architecture into the embedding layers, prototype learning layer and output layer, as shown in Figure \ref{fig:diagram}. We extend the prototype network to extract channel-specific prototypical patterns in geoscientific multi-channel image based datasets. Information on the specific notation used in the study is provided in Table \ref{notation_table}.

\begin{table}[tb!]
\tabcolsep=0pt%
\TBL{\caption{Notation Glossary\label{tab3}}}
{\begin{fntable}
\def\arraystretch{1.5}
\begin{tabular*}{\textwidth}{@{\extracolsep{\fill}}lcccccc@{}}\toprule%

 \TCH{Definition} & \TCH{Symbol} & \TCH{Dimension}\\\midrule

{\TCH{Number of Channels}}                      &   $C$             &  $\mathbb{N}$\\
{\TCH{Number of Output Classes}}                &   $K$             &  $\mathbb{N}$\\
{\TCH{Number of Prototypes per Channel $j$}}    & $N$               &  $\mathbb{N}$\\
{\TCH{Number of Total Prototypes }}             & $A = N \cdot C$   &  $\mathbb{N}$\\
{\TCH{Embedding Dimension}}                     & $d$               &  $\mathbb{N}$\\

\hline

{\TCH{Training Sample $i$}}             &  $X^{(i)} = \left[X^{(i)}_1 \,... \, X^{(i)}_C\right]$     &  $\mathbb{R}^{H \times  W \times  C}$\\
{\TCH{$j^{th}$ channel of $X_{i}$}}     &   $X^{(i)}_j$                             &  $\mathbb{R}^{H \times W }$\\

\hline

{\TCH{Encoder}}                             & $E(\text{input; weights})$        &  $E: \mathbb{R}^{H \times W } \mapsto \mathbb{R}^{h \times w \times d}$\\
{\TCH{Prototype Layer for Channel $j$}}     & $P_j(\text{embedding; weights})$  &  $P: \mathbb{R}^{h \times w \times d } \mapsto \mathbb{R}^{N \times h \times w}$\\
{\TCH{Output Linear Layer}}                 & $L(\text{sim. values; weights})$  & $L: \mathbb{R}^{A} \mapsto \mathbb{R}^{K}$ \\

\hline

{\TCH{Trainable Encoder Weights}}               & $\theta_E$        & varies \\
{\TCH{Trainable Prototypes for Channel $j$ }}   & $\phi_j$          &  $ \mathbb{R}^{N \times d}$\\
{\TCH{Trainable Location Scaling Parameter}}    & $\omega_j$        &  $ \mathbb{R}^{N \times h \times w}$\\
{\TCH{Trainable Linear Layer Weights}}          & $\psi_L$          &  $ \mathbb{R}^{A \times K}$\\

\hline

{\TCH{$X^{(i)}_j$ Embedding of Channel $j$}}        & $Z_j^{(i)} = E (X_j^{(i)}; \theta_E)$                 & $ \mathbb{R}^{h \times  w \times  d}$ \\
{\TCH{$X_j^{(i)}$ Prototype Similarity Grid}}       & $S_j^{(i)} = P_j (Z_j^{(i)}; \phi_j)$                 &  $ \mathbb{R}_{\ge 0}^{N \times h \times w}$\\
{\TCH{$X_j^{(i)}$ Maximum Similarity Values}}       & $M_j^{(i)} =   \displaystyle\max S_j^{(i)}$           &  $ \mathbb{R}_{\ge 0}^{N}$\\
{\TCH{$X^{(i)}$ Concat. Maximum Similarity Values}} & $M^{(i)} = \text{concat}(M_1^{(i)}...M_C^{(i)})$      &  $ \mathbb{R}_{\ge 0}^{A}$\\
{\TCH{Softmax Output for Sample $X^{(i)}$}}         & $y^{(i)} = \text{softmax}\left(L(M^{(i)}; \psi_L)\right)$        &  $ \mathbb{R}^{K} \in [0,1]$\\

\botrule
\end{tabular*}%
\end{fntable}}
\label{notation_table}
\end{table}

\subsubsection{Encoder} 
As in standard neural networks, the encoder consists of a set of layers (2D convolution layers, average 2D pooling layers and activation layers) that extract relevant features from the input data. The number of individual convolution, pooling and activation layers is flexible and varies dependent on the particular learning task.  A key advantage of the encoder is its flexibility in the combination of neural network layers used to produce an embedding, though larger embedding spaces can increase the prototype search time and the projection step, detailed in Section \ref{projection}. 

The training data consists of samples each belonging to one of $K$ output classes. One sample is a singular multi-channel image, $X^{(i)} = [X^{(i)}_1 \, ...  \, X^{(i)}_C] \in \mathbb{R}^{H \times  W \times  C}$, where $H$ represents the height, $W$ represents the width and $C$ represents the number of channels. The training data is separated by channel where $X^{(i)}_j$ represents channel $j$  of the $i$th training sample, $X^{(i)}$. Similar to \citet{ghosal2021multi}, each channel, $X^{(i)}_j$, is separately processed to produce a channel-specific embedding, $Z_j$. Unlike \citet{ghosal2021multi}, which uses a separate encoder for each channel, we employ a shared encoder, $E$, with parameters $\theta_E$, reducing model size and training complexity without loss in model performance (see Supplement Section 1). The embedding of channel $j$ of sample $X_j^{(i)}$ is shown in Equation \ref{encoder} with encoder $E:\mathbb{R}^{H \times W } \mapsto \mathbb{R}^{h \times w \times d}$ where the dimension of the embedding, ($h$, $w$ and $d$) will depend on the architecture of the encoder.
\begin{equation}
    Z_j^{(i)} = E (X_j^{(i)}, \theta_E)
\end{equation}
\label{encoder}

\subsubsection{Prototype Learning Layers}

We introduce prototype learning layers that learn the representative prototypical patterns for each channel and each class, allowing the model to compare the new inputs against the library of learnt example prototypes to make a prediction. The sample's embedding for each channel, $Z_j^{(i)}$, is subsequently fed into its own corresponding separate prototype learning layer $P_j$ where the trainable parameters, $\phi_j$, correspond to $N$ prototypes, each of dimension $d$ in the embedding space. For each class label $k$, the model learns $N$ prototypes per channel for a total of $A = N \cdot C$ prototypes per class.

Following \citet{chen2019looks}, the model computes a function of the $L2$ distance between the embedding, $Z^{(i)}_{j}$, and each individual $\phi_{j}^{(n)}$ prototype for $n= 1 ...N$ within channel $j$, producing a distance map of shape $\mathbb{R}^{h \times w \times N}$. \citet{chen2019looks} apply this function across all channels, while we apply this function independently in a channel-wise manner. Taking approximately the inverse of the distance map in Equation \ref{sim_equation} produces a similarity grid, $S^{(i)}_{j}$, for each prototype, where higher similarity values indicate closer matches to that prototype. Depending on the learning task, we include an optional location scaling grid parameter, $\omega_j$, assigned to each prototype, \citep{barnes2022looks} which is an additional trainable parameter of identical dimensions to the similarity grid, $ \mathbb{R}^{N \times h \times w}$, that weights similarities differently based on their location in the grid. The resulting similarity map for each prototype is multiplied with their respective location scaling grid. The details of these modifications are discussed further in the three case studies below. 
 \begin{equation} \label{sim_equation}
    S_j^{(i)} = P_j \left(Z_j^{(i)}; \phi_j\right) = \log\left(\frac{||Z_j^{(i)}-\phi_j^{(n)}||_2^2+1}{||Z_j^{(i)}-\phi_j^{(n)}||_2^2 + \epsilon}\right) \cdot \omega_j, \quad n=1  ... N
\end{equation} 
We take the singular maximum value, $M_j^{(i)}$, (a 1 $\times$ 1 patch) from each similarity grid, $S_j^{(i)}$ for each prototype within channel $j$ i.e the value at the point where the channel prototype and embedding is most similar. We concatenate the $N$ maximum similarity values across each channel into $M^{(i)}$ resultins in $A = N\cdot C$ similarity values for each sample, $X^{(i)}$, with each element representing how similar the input sample is to the corresponding learnt prototypes across all channels.
\begin{equation}
    M_j^{(i)} =   \displaystyle\max_{1\leq u \leq h, 1\leq v \leq w} S_j^{(i)} [u,v,:] , \quad \mathbb{R}^{h\times w \times N } \mapsto \mathbb{R}^{N}
\end{equation}

\subsubsection{Output Layer} 

The concatenated maximum similarity values are fed into the final softmax linear classifier, $L$ with trainable parameters, $\psi_L$. $L$ consists of a linear layer with zero bias and softmax activation to produce output $y^{(i)}$. The model prediction is determined from a weighted combination of all of the computed prototype similarity values in Equation \ref{classifier_eqn}. Softmax is applied to compute logit probabilities for each class $k$ and the class with the maximum probability is chosen. We randomly initialize this layer to allow the model to adequately learn which prototypes from specific channels may be important for specific classes as prototypes from different channels could have varying relevance levels (including zero relevance) to the learning task. 
\begin{equation} \label{classifier_eqn}
    y^{(i)} = \text{softmax}(L(M^{(i)}; \psi_L)), \quad\mathbb{R}^{A} \mapsto \mathbb{R}^{K} \in [0,1]
\end{equation}

\subsection{Training Procedure}

\subsubsection{Three Stage Training}
The model is trained using the three-stage approach in \citet{ barnes2022looks}. In stage 1, the model learns the encoder's parameters, $\theta_E$, the prototypes, $\phi$ and, if included, the location scaling parameters, $\omega$ while the final layer weights, $\psi_L$, are frozen. Stage 2 is a periodic ``projection step'' (detailed below in Section \ref{projection}) that constrains prototypes to correspond to latent patches from specific training instances by projecting each prototype onto its most similar training embedding, enabling meaningful prototype visualization. In stage 3, all layers except for the final classification layer are frozen, and the prediction weights, $\psi_L$, are learnt. Within a training cycle, stages 1 and 3 are run sequentially, while the projection step is applied at task-dependent intervals, a tunable hyperparameter. Dropout is used during stage 1 to reduce overfitting but is disabled prior to projection to ensure stable embeddings.

\subsubsection{Loss Functions}
In Stage 1, we use a loss function consisting of the sum of the standard cross entropy term and three channel-specific prototype loss terms. Following the approach in \citet{chen2019looks}, we include (1) a cluster loss which ensures the prototypes are close to patterns of the correct class and (2) a separation loss which ensures the prototypes are far from patterns of incorrect classes. We also include (3) a diversity loss \citep{ming2019interpretable} which ensures prototypes are distinct from each other. In the diversity loss, a threshold parameter is used which sets the allowed minimum distance between prototypes. For each channel, we compute separate cluster, separation and diversity losses. 

In stage 3, we use a standard cross-entropy loss to fine-tune the classifier weights and include $L1$-regularization on $\psi_{L}$ to create a sparser weight matrix to encourage a model where fewer prototypes contribute to the final classification for improved interpretability.

\subsubsection{Prototype Projection Step} \label{projection}

The stage 2 projection step ensures prototypes remain interpretable and visualizable by replacing the learned prototype parameter values with similar patches selected from training examples that we can visualize. To do this, we select a $1\times 1 \times d$ patch (a single point in the embedding space) from a training example to update the prototype. The training example is chosen such that the $L2$ distance between the prototype and the selected patch is minimized following Equation \ref{sim_equation}. The projection step is performed separately for each channel and occurs every few training cycles to allow for continuous model training of the prototypes before replacement. The frequency of projection is a hyperparameter set by the user.

\subsection{Generating Explanations}

To understand the model decision-making process, we generate explanations by leveraging the prototype structure built into the model architecture. We use this structure to examine which prototypical patterns in different channels provide evidence for predictions at the instance-level as well as for model overall reasoning across the dataset. 

\subsubsection{Local Explanations}

A local explanation provides evidence for the model prediction associated with a specific data instance (``test sample''). For a given test sample, $X^{(i)}$, we quantify how much each prototype contributed to the prediction for a particular class with $S^{(i)}$, also referred to the \textit{prototype score}. This contribution score is calculated by multiplying the maximum similarity value, $M_j^{(i)}$, between the test sample and each prototype, $\phi_j$ by the corresponding learned weight within $\psi_L$ that links that prototype to the predicted class. We rank the prototype contribution scores to identify the highest scoring prototypes, which correspond to patches in the embedded representations of specific instances in the training dataset that are most similar (within the embedding space) to the test sample. Each prototype is associated with a particular data channel and its contribution highlights the channel-specific patterns that influenced the prediction of the test sample. High-scoring prototypes representative of a specific class that show high similarity to the test sample provide the instance-based evidence underlying the model’s prediction. 

We can visualize the prototypes using their \textit{receptive field} which shows the set of pixels in the original image space that correspond to the patch location in the embedding space \citep{araujo2019computing}. To compute the receptive field, we follow a brute-force ``ping'' method used in \citet{barnes2022looks} to map the prototypical pattern represented in the original image space. By ensuring prototypes are derived from actual instances in the training data, the neural network reasoning is easier to interpret.

\subsubsection{Global Explanations}

Global explanations can provide insight into the overall model decision making process, for broader patterns in the data set, rather than for a single instance, For example, global explanations might reveal the degree to which specific features or channels are relevant to each class label. To generate global explanations, we can analyze results such as the aggregate frequency of high scoring prototypes over the entire test set, $\mathcal{X}$ and can inspect the variance in model assigned prototype weights, $\psi_L$, across specific classes and channels.

One type of global explanation, relative channel importance, can be derived from the final classification weight matrix, $\psi_L$. Since the model decision is based on a weighted linear combination of the prototype similarity values for each class, these weights can indicate the overall level of importance of prototypes from a specific channel for each class. Examples of this type of global explanation is shown in Section \ref{sec:MNIST} Figure \ref{mnist_finalweights} and in Section \ref{sec:MJO} Figure \ref{fig:mjo_noise_weights}. A larger range of positive and/or negative weights assigned for prototypes from a specific channel would indicate a larger influence of those channel prototypes for the task. On the other hand, prototypes from a channel assigned near zero weights in $\psi_L$ would indicate that those prototypes do not contribute to model predictions (i.e.\ this channel is not relevant to the task).

Prototypes themselves can provide global level insight into the learning task. The prototypes, $\phi$, are learnt in a way such that they are representative of each class. We can compute and rank the individual prototype scores for all instances across the test set, and then identify the prototypes that are most frequently high scoring for a particular class, $k$. An example of this type of global explanation is shown in Section \ref{sec:EO} Figure \ref{fig:rs_global}. Channels that provide more of the high scoring prototypes suggest that information within these channels is more relevant to the specific class and channels that provide less high scoring prototypes contain less relevant information to the class, similar to various feature importance method analyses traditionally used in machine learning methods. 

\section{Case Studies}

In the following three sections, we detail the datasets used in the three case studies. We develop a synthetic dataset (Section \ref{sec:MNIST}) for the first case study to showcase an example of how the model works and the explanations it provides. We then demonstrate the method on two multi-channel geoscientfic classification tasks: one using gridded climate data (Section \ref{sec:MJO}) and the other remote sensing imagery (Section \ref{sec:EO}). For each case study, in addition to our proposed method, we train a standard neural network with an identical encoder architecture and a standard linear classifier layer as a baseline for comparison with our prototype-based architecture. We also compare our channel-specific prototype method against a standard prototype network following \citet{barnes2022looks}), where channels processed together and prototypes are learnt jointly across channels rather than individually, referred to in the results as the Prototype Network (No Channel-Specific Prototypes). Details for the model parameters for each of the case studies can be found in Table \ref{tab2}. Within each section, we include the Data and case study task, Model architecture and training, Results, and Discussion with a final overall discussion in Section \ref{Discussion}.

\begin{table}[tb!]
\tabcolsep=0pt%
\TBL{\caption{Model parameters for three case studies discussed in study. Case Study 1 refers to the synthetic MNIST Case study in Section \ref{sec:MNIST}. Case Study 2 refers to the Madden-Julian Oscillation Case study in Section \ref{sec:MJO}. Case Study 3 refers to the Land Cover Classification case study in Section \ref{sec:EO}. \label{tab2}}}
{\begin{fntable}
\begin{tabular*}{\textwidth}{@{\extracolsep{\fill}}lcccccc@{}}\toprule%

\TCH{Model Training Parameters} & \TCH{Case Study 1} & Case Study 2 & Case Study 3\\\midrule

{\TCH{Number of class labels $(K)$}}                     &10            & 9                 & 10  \\
{\TCH{Number of Channels $(C)$}}                         & 3            & 3                 & 13 \\
\TCH{Batch size}                                         & 32           &32                 & 64\\
{\TCH{Learning Rate }}                                   & 0.001        & 0.001             & 0.001 \\
{\TCH{Frequency of Prototype Projection}}                & 3 cycles     & 3 cycles          & 2 cycles \\
{\TCH{Encoder Layer Output Channels}}                    & 8, 16, 32    &16, 32, 64, 64     & 16, 32, 64, 64 \\
{\TCH{Embedding Dimension (h $\times$ w $\times$ d)}}    & 2 $\times$ 2 $\times$ 32  & 2 $\times$ 5 $\times$ 64 & 2 $\times$ 2 $\times$ 64  \\
{\TCH{Number of Prototypes per Class per Channel $(N)$}} &  5           & 10                & 4 \\
{\TCH{Number of Total Prototypes $(N\cdot C \cdot K)$}}  &  150         & 270               & 520 \\
{\TCH{$L1$ Regularization Coefficient }}                  & 0.01         & 0.001             & 0.001 \\
{\TCH{Cluster Cost Coefficient}}                          & 0.7          & 0.5               & 0.2 \\
{\TCH{Separation Cost Coefficient}}                       & 0.7          & 0.2               & 0.02 \\
{\TCH{Diversity Cost Coefficient}}                        & 0.001        & 0.01              & 0.01 \\
{\TCH{Diversity Cost Threshold}}                         & 0.001        & 0.001             & 0.1 \\

\botrule
\end{tabular*}%
\end{fntable}}
\end{table}

\section{Synthetic MNIST Case Study}\label{sec:MNIST}
\subsection{Dataset and Task}
We use a synthetic case study to demonstrate the ability of our proposed method to make predictions that are accurate and generate explanations consistent with the expected channel-specific reasoning of the model. To create the data set for this task, we modify the MNIST (Modified National Institute of Standards and Technology) greyscale hand-written digit dataset commonly used as a benchmark for image classification tasks and introduce an known, channel-dependent rule to generate class labels. 

We construct $56\times56\times3$ images divided into four $28\times28$ quadrants (the size of original MNIST data) with three channels. To create the synthetic image, we sample a ``side", right or left (each with probability $0.50$ ) and draw three random digits from the MNIST dataset; a free digit (0 to 9), an even digit (0, 2, 4, 6 or 8) and an odd digit (1, 3, 5, 7, or 9). The free digit is placed in all four quadrants of channel 1. The even and odd digits are placed in channels 2 and 3 respectively, but only in the two quadrants according to the the selected ``side" (right or left). See Panel (a) in Figure \ref{mnist_98} for an example of a synthetically generated image. We assign labels to the MNIST-derived synthetic images using the following rule: the even digit in channel 2 is used as the label when the ``right" side is selected, and the odd digit in channel 3 is used when the ``left" side is selected.

The task is designed such that there is important channel-specific information for the model to decipher. First, the digit in channel 1 holds no information relevant to the sample's label. For even numbered labels, the relevant number patterns are located only in channel 2 on the right hand side. For odd numbered labels, the relevant number patterns are located only in channel 3 on the left hand side. Consequently, for even numbered labels, the ``blank" digit patterns (the quadrant without a digit pattern) are only located on the left hand side of the image and vice versa for odd numbered labels. 

\subsection{Model Architecture and Training}

We generate 12000 samples: 8640 for training, 2160 for validation, and 1200 for testing. Each $56\times56$ channel is separated and individually fed into an encoder consisting of three sequences of convolution, leaky ReLU activation and average pooling layers. For the ten available classes, we learn five prototypes per class, resulting in 50 prototypes for each channel and 150 prototypes in total. Each prototype is associated with a learned location-scaling parameter that identifies the spatial regions where the model assigns the greatest weight to similarity with the prototype. We include this parameter due to the location-dependent (i.e the chosen ``side") relevance of both digit and blank patterns for each label. For this task, the stage two prototype projection step is completed every 6 epochs (3 cycles of stage 1 and stage 3 training). Column 1 in Table \ref{tab2} outlines the model training parameters. Model hyperparameters were minimally tuned manually.

\subsection{Results}
\subsubsection{Model Performance}
The MNIST classification prototype based network with the channel specific prototypes achieves an overall accuracy of $0.978$ and individual class accuracies ranging from 0.950 to 1.0. In comparison, the standard neural network (with no prototype layer) with identical model architecture and training achieves an accuracy of $0.937$ as shown in Table \ref{tab_minst}. In comparison, a standard prototype network without channel-specific prototypes, trained for the same number of epochs, only achieves an accuracy of $0.685$. The synthetic task's labeling rule is explicitly channel-dependent, so we expect the standard prototype network with jointly learnt prototypes to underperfom the channel-specific method, highlighting the importance of channel-specific prototype explanations.

\begin{table}[htb!]
\tabcolsep=0pt%
\TBL{\caption{Case Study 1: Synthetic MNIST Classification Performance\label{tab_mnist}}}
{\begin{fntable}
\begin{tabular*}{\textwidth}{@{\extracolsep{\fill}}lcccccc@{}}\toprule%

\TCH{Model} & \TCH{Classification Accuracy} \\\midrule

{\TCH{Standard Neural Network}}&  0.937\\
{\TCH{Prototype Network (No Channel-Specific Prototypes)}}&  0.685\\
\TCH{Prototype Network with Channel-specific Prototypes }& 0.978\\

\botrule
\end{tabular*}%
\end{fntable}}\label{tab_minst}

\end{table}
\subsubsection{Local Explanations}
\begin{figure}[htb!]
\centering
\includegraphics[width=0.95\linewidth]{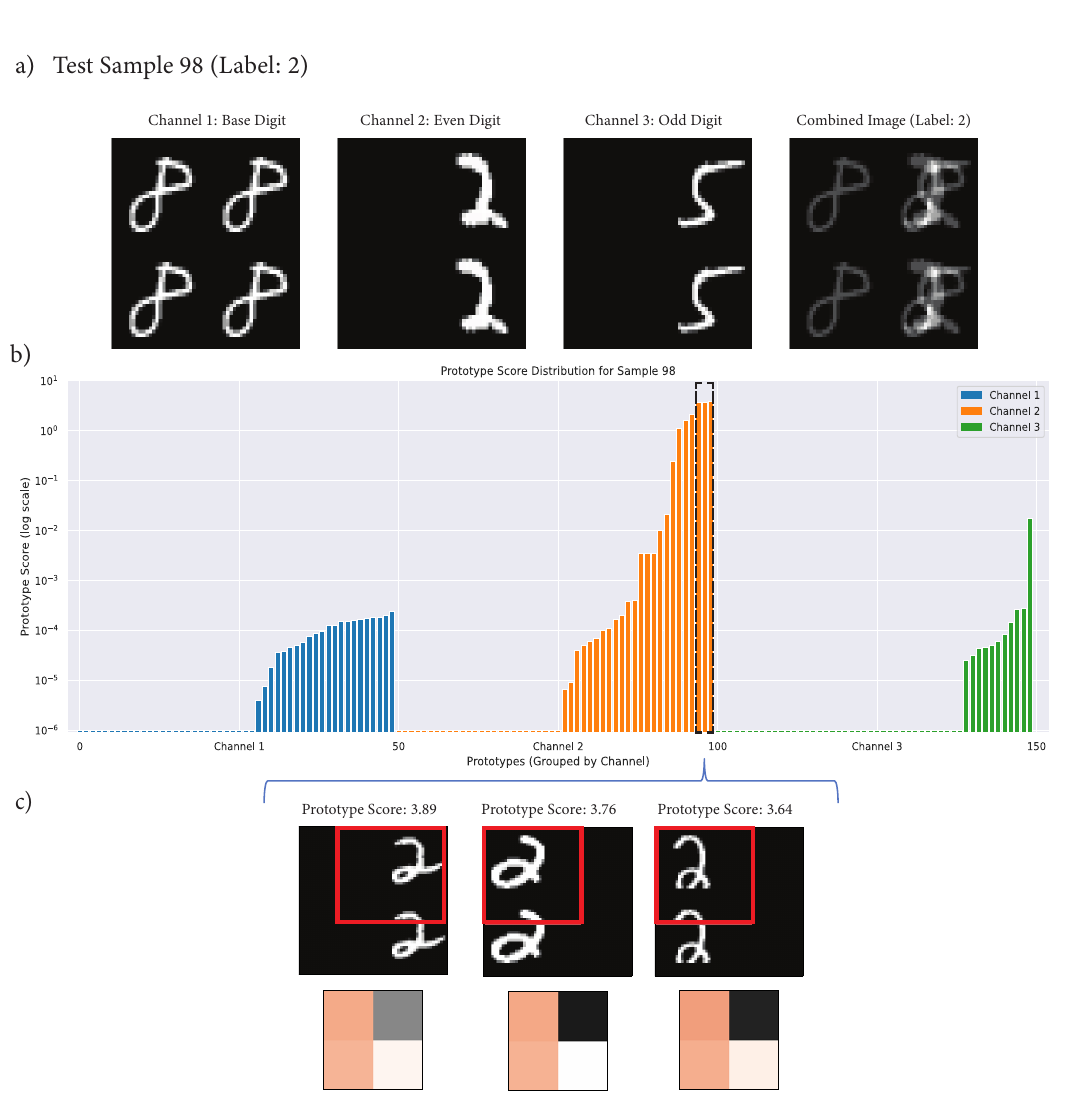}
\caption{ (a) Example of synthetically generated sample. The left three images show three image channels separated. Image on right shows the sample with channels concatenated together (channels shown stacked for visualization purposes). Ground truth label: 2 (b) Test Sample 98's prototype score distribution (on logarithmic scale, capped below at $10^{-6}$) for prediction of class ``2''. Prototype scores are grouped by channel where blue, orange and green bars represent Channel 1, 2, and 3 prototypes respectively and sorted in ascending order. The dashed outline box highlights the top three scoring prototypes that contributed to this prediction visualized in Panel c. (c) Top three scoring prototypes shown within the red outline boxes capturing ``2'' digit pattern displayed along with the original training image from which the prototype originates. Below each prototype is its associated location scaling  indicating ``side'' where similarity to the prototype is weighted strongly (red indicates negative relevance, grey indicates positive relevance). \\ \textit{\textbf{Explanation of model prediction:} The model predicts that the image belongs to class ``2'' because of a high similarity to the ``2'' digit pattern in the right side of channel 2 in training image 8613, which makes the largest contribution to the final prediction.}}
\label{fig:mnist_98}
\end{figure}

For instance-level explanations, we expect our approach to identify the channel-specific pattern that determines the label: digit information in channel 2 (even label) or channel 3 (odd label) according to their placement on the ``right'' or ``left'' side, respectively. We present a representative example of the local channel-specific explanations in Figure \ref{fig:mnist_98}. The topmost panel shows test sample $98$ contains digits ``8'', ``2'' and ``5'' across its three respective channels. Under the labeling scheme, this sample belongs to class $2$ as digits appear on the right side. Accordingly, the highest scoring prototype should match the ``2'' digit pattern on the right hand side of channel 2, as channels 1 and 3 do not influence the ground truth class label. 

We show the distribution of the 150 prototype scores contributing to the class $2$ prediction grouped by channel and sorted in ascending order within the middle panel of Figure \ref{fig:mnist_98}. The distribution reveals that most of the top scoring prototypes belonged to channel 2. We highlight the top three scoring prototypes, each belonging to channel 2 shown within the red-outline boxes in the bottom panel of Figure \ref{fig:mnist_98}. For visual clarity, we display the receptive field of the prototypes, since prototypes reside in the latent space and we also show the full training images from which the prototypes originate, even though these larger images do not influence model decisions. Below each prototype, we show the associated location scaling grid which indicates where the similarity is most strongly weighted in the latent space (i.e.\ the right side or the left side), not the location of the prototype within the larger training image. 

As we expect, the highest scoring prototype (score $3.890$) (leftmost in bottom panel) corresponds to the ``2'' digit pattern from training sample $8613$. The associated location scaling grid for this prototype assigns positive relevance to the right hand side as indicated by the grey shading and negative relevance to the left hand side as indicated by the red shading. The distinction between relevance of the top and bottom quadrants within the location scaling is arbitrary for this task. The second and third highest scoring prototypes shown in the middle and rightmost figures of the bottom panel, similarly identify the ``2'' digit pattern from different training examples. Prototypes in the other channels have negligible contribution scores, consistent with their irrelevance for this label, as seen by the highest green and blue scores in Panel b. Overall, the model correctly identifies relevant digit patterns in channel 2 as the key contributor for this prediction. In contrast, the prototype network without channel-specific prototypes misclassifies this sample because it fails to find a jointly learned prototype similar to the input sample across all channels (see Figure S1 in Supplement Section 2) highlighting the need to identify channel-specific patterns in channel-dependent learning tasks.

\subsubsection{Global Explanations}

\begin{figure}[htb!]
\FIG{\includegraphics[width=1\textwidth]{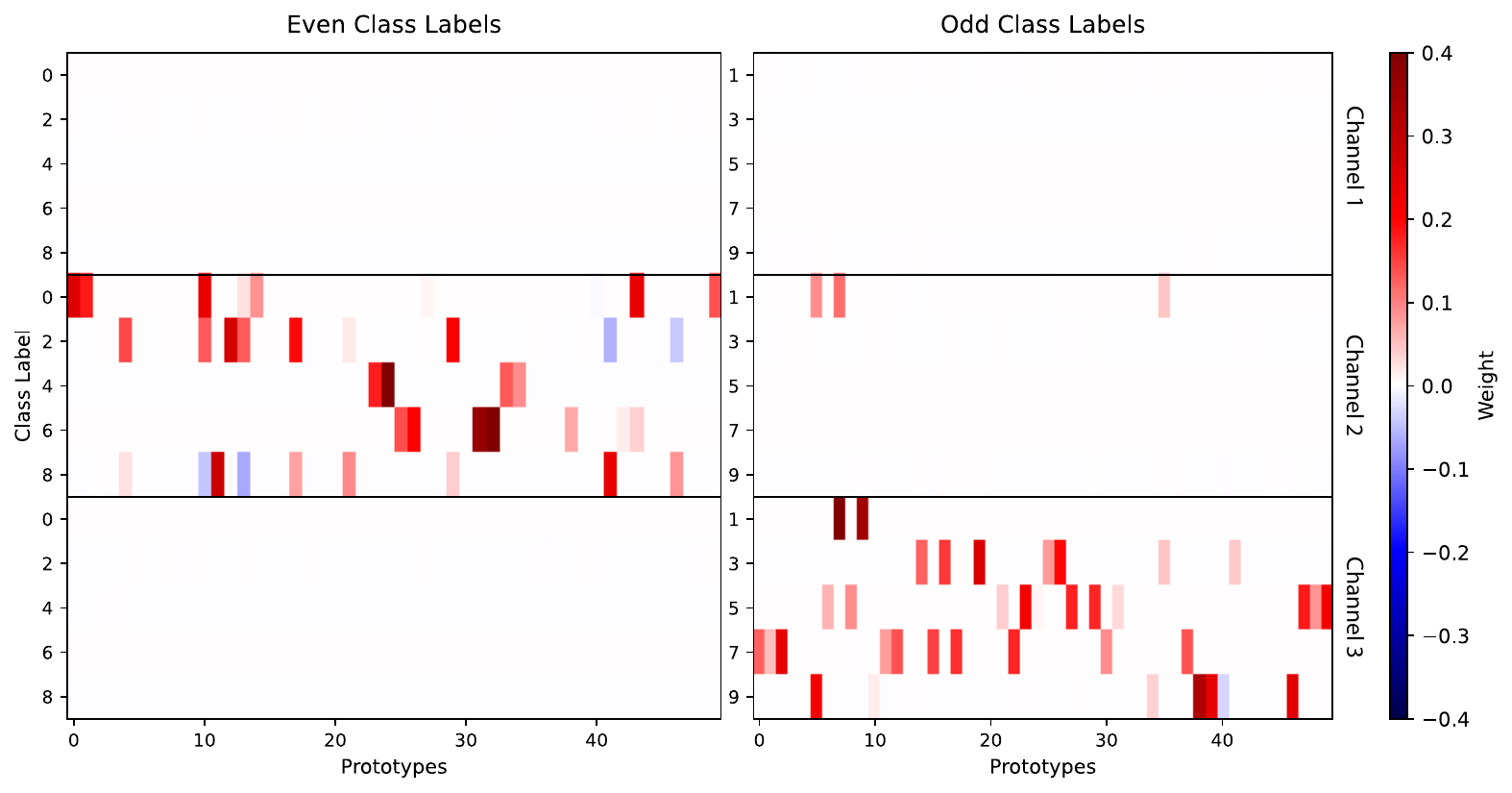}}
{\caption{MNIST Case Study final linear layer weight matrix (rows and columns permuted for visual clarity) separated into even digit class labels (0,2,4,6,8; left panel) and odd digit class labels (1,3,5,7,9; right panel). The matrix shows the weights associated with the 50 prototypes from each channel (top: channel 1, middle: channel 2, bottom: channel 3). Strong positive (red) and negative (blue) weights are predominantly assigned to channel 2 for even class labels and to channel 3 for odd class labels while near zero (white) weights appear for prototypes from channels that are irrelevant to the corresponding class.}
\label{mnist_finalweights}}
\end{figure}

At the global-level, inspecting the final linear layer weight matrix, in Figure \ref{mnist_finalweights}, reveals that the model is using the expected class and channel relationships to make predictions across the test set. When aggregating prototype similarities for the final softmax classification, for even numbered classes, channel 2 prototypes are assigned higher weights and channel 3 prototypes are assigned lower, near zero weights. We see the opposite for odd numbered classes; channel 2 prototypes are assigned lower, near zero weights and channel 3 prototypes are assigned higher weights. Most channel 1 prototypes are assigned near zero weights and do not exhibit a large range of positive or negative values, following the channel-dependent labeling rule used to design the synthetic task. 

\begin{figure}[htb!]%
\FIG{\includegraphics[width=1\textwidth]{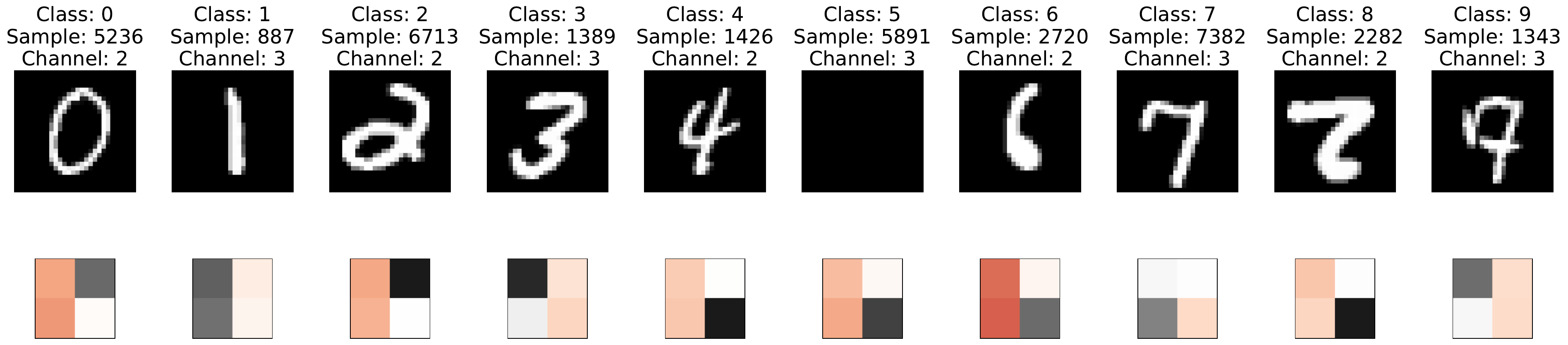}}
{\caption{In the top row, across the test set, the most frequent high scoring model-identified prototypes (digit or ``blank'' pattern) are shown for each class (0-9). Information on the specific channel and training sample from which the prototype originates from is included. Bottom row shows each prototype's associated location scaling grid which shows the 'side' (left or right) of an input sample where the prototype pattern is more relevant where gray shading indicates positive relevance and red shading indicates negative relevance. For even-numbered classes, the prototype digit pattern is more relevant on the right hand side and for vice versa for odd-numbered classes, with the exception of class label 5 due to the ``blank'' prototype identified.}
\label{mnist_freqproto}}
\end{figure}

Across the correctly classified test samples within a specific class, we can inspect the most frequent high scoring prototypes as global insight into those prototypes that most commonly contribute to the specific class prediction. In Figure \ref{mnist_freqproto}, the most frequent highest-scoring prototype for each class is displayed alongside its associated location scaling grid. For even numbered classes (0,2,4,6,8) the model identifies the corresponding matching digit prototype pattern as the most frequent high scoring prototype, with the exception of class label 8. For class label 8, the prototype incorrectly captures the `2' digit pattern relevant on the right side; this may be due to similarity between `2' and `8' digit patterns. These even-number class prototypes belong to channel 2 of the original training samples from which they originate. The associated location grids reveal that these prototype patterns are more relevant to their respective classes when located in the right hand side of the test sample. This agrees with our method used to develop the synthetical samples for the even-numbered class labels.

With the exception of class label 5, the odd numbered classes (1,3,7,9), the model identifies the corresponding matching digit pattern as the most frequent high scoring prototype. The location grids reveal these prototype patterns are more relevant when located in the left side of a test image, opposite to what we saw for the even numbered class labels. For class label number 5, a ``blank'' digit prototype was identified (the prototype corresponds to the empty pixels within the training image). We expect blank digit patterns to also be relevant for this learning task since the absence of a digit in the right hand side of the image is a characteristic of odd numbered labels. Additionally, we note that none of these prototypes contained patterns from channel 1 which did not have relevant information for the task, indicating the model correctly identifies highly similar prototypes from the corresponding channels in order to make predictions for this learning task.

\subsection{Case Study Discussion}
Our results demonstrate that the channel-specific prototype model reliably identifies the decisive channel-specific patterns required by the synthetic task: even digits on the right side of channel 2 for even classes and odd digits on the left side of channel 3 for odd classes. For an instance-based explanation, when two of the three channels are irrelevant to a prediction, their similarity scores to corresponding prototypes are lower or are assigned suppressed weights in the final linear layer, ensuring they contribute minimally to the decision. In contrast, standard prototype networks with jointly learned prototypes can fail to find an appropriate matching prototype across all channels. Moreover, the synthetic dataset comprises of a large combinatorial space with 10 digits, even-odd structure and left-right spatial placement. Capturing all combinations with shared prototypes present a substantial challenge for standard prototype methods as individual prototypes are forced to simultaneously represent multiple diverse patterns across channels. In contrast, maintaining separate channel prototypes allows each prototype to independently capture the relevant patterns while ignoring irrelevant patterns from other channels. 

Global explanations further show the model learns the expected channel and class relationships governing the task, assigning higher weights to relevant channel prototypes and suppressing those that contain less relevant patterns. Across the test set, the model frequently highlights the correct digit or blank prototype patterns relevant to each class. Overall, these findings demonstrate channel-specific prototypes provide both instance-level and global-level explanations while maintaining predictive performance relative to a standard neural network and a prototype network without channel-specific prototypes for a explicit channel-dependent learning task.

\section{Madden Julian Oscillation Phase Classification Case Study}\label{sec:MJO}
\subsection{Dataset and Task}
We demonstrate our method on a geoscientific use-case: a climate classification task following the Madden Julian Oscillation (MJO) phase \citep{madden1994observations} classification case study in \citet{barnes2022looks}. The MJO is an eastward propagating convective pattern over the Indian and Pacific Oceans on sub-seasonal timescales, influencing phenomena such as monsoon onset and intensity \citep{zhang2005madden}. It consists of eight active phases, defined by location and strength of convection, with convection initiating near eastern Africa in Phase 1 and propagating eastward into the Pacific (Phase 8) \citep{knutson198730}, and one inactive phase (Phase 0). Each phase has defining characteristics across multiple environmental variables lending itself to be an appropriate use-case for our channel-specific prototypical method to test how well a model can identify MJO phase from current convective patterns and atmospheric state variables. 

In this case study, we use three environmental input variables (channels): daily outgoing longwave radiation (OLR), zonal wind at 200hPa (U200) and zonal wind at 850hPa (U850) data \citep{barnes2022looks} and \citep{toms2020testing_code} from January 1 1980 to December 31 2016 over the tropical region (15N to 15S degrees latitude, 0 to 260E degrees longitude, 2$^{\circ}$ $\times$ 2$^{\circ}$ resolution) sourced from NOAA once-daily OLR climate data record \citep{lee2014noaa} and NASA MERRA-2 reanalysis \citep{gelaro2017modern}. OLR reflects convective activity and cloud cover, while zonal wind describes upper and lower level wind patterns over the region. \citet{barnes2022looks} employed prototypes that consisted of a joint pattern across all three channels. However, MJO phases can be characterized by spatial patterns that differ by variable and location. For example, in Phases 1 and 2, OLR shows active convection over eastern Africa, where U200 diverges at the top of the atmosphere and U850 converges near the surface \citep{wheeler2004all} with an opposite dipole with reversed wind patterns over the Indian ocean \citep{madden1972description}. Channel-specific prototypes can help address this by allowing each variable to contribute its own prototypical spatial pattern, making it better suited for tasks where relevant information can vary across channels and regions. 

We follow a similar data processing set-up to the deep learning-based approach of \citet{toms2021testing}, using precomputed MJO phase labels derived from the Outgoing Longwave Radiation MJO Index (OMI; \citep{kiladis2014comparison}). Following \citet{toms2021testing}, the OMI phases 0-8 are defined by the azimuth of the two leading principal components obtained from projections of 20-96 day filtered OLR anomalies onto the OMI phase space and subsequently aligned with the Real-Time Multivariate MJO Index Series (RMM) phase space \citep{wheeler2004all}. For detailed information on the OMI construction and phase-space labeling, we refer to \citet{toms2021testing}. The MJO phase labels are fixed prior to model training and no phase-space projection or temporal filtering is performed within the model learning framework. While \citet{toms2021testing} exclude inactive MJO phase periods from their dataset, we retain inactive Phase 0 in our analysis, following \citet{barnes2022looks}'s definition where the amplitude of the MJO is below 0.5 in the phase space.

\subsection{Model Architecture and Training}
Each gridded image in the dataset has the dimensions $16 \times 131 \times 3$ (latitude $\times$ longitude $\times$ channels). We follow \citet{barnes2022looks}'s random selection process and reserve data from three randomly selected years ($1984, 1987, \text{and } 2015$) for validation and three years ($1989, 2008, \text{and } 2011$) for testing, and the remaining 31 years for training. The dataset is split with 11315 samples for training, 1095 samples for validation and 1095 samples for testing, where leap days are excluded for consistency with \citet{toms2021testing}. Each channel is split and individually fed into the shared encoder consisting of four sequences of convolution, leaky ReLu activation and average pooling layers. For the 9 available classes, we aim to find 10 prototypes per class resulting in 90 prototypes for each channel and 270 prototypes in total. Following \citet{barnes2022looks}, we include the location scaling parameter since the relevance of spatial patterns associated with MJO phases are location dependent. The stage 2 prototype projection step is completed every 10 epochs or 5 cycles of stage 1 and stage 3 training. After multiple cycles of stages 1 and 3, once the prototypes are learnt and finalized, the model trains in stage 3 for 15 extra epochs in order to finalize prototype weights. Column 2 in Table \ref{tab2} outlines the model training parameters. Model hyperparameters were tuned using the Weights and Biases sweeps feature \citep{wandb}.
\subsection{Results}
\subsubsection{Model Performance}
For the MJO classification task, the model with channel-specific prototypes performs with overall $0.553$ test accuracy.  As the MJO cycle is a continuous phenomenon, the class boundaries are not rigid so we also report the $+1/-1$ class accuracy of $0.805$ (indicating accuracy within one MJO Phase of the correct label). The individual class accuracies range from $0.41$ to $0.73$ (see Figure S2 in Supplement Section 3), with most errors due to the model predicting a phase before or after the correct phase -- pointing to the continuous nature of the MJO cycle with ambiguous Phase boundaries. Another source of incorrect predictions are false negatives: the model misclassifying active phases as the inactive Phase 0. The inactive phase is determined by the intensity of the MJO pattern rather than the location of convection and zonal wind patterns indicating the model can correctly identify the eastward movement of prototypical patterns across phases but struggles in capturing prototypical patterns relating to the intensity of the MJO. 

Though the overall class accuracy is not very strong, it is comparable to a standard neural network's accuracy of $0.525$ trained on the same data with the same encoder architecture. A prototype network that learns prototypes jointly across channels (without channel-specific prototypes, following \citet{barnes2022looks}) achieves an accuracy of $0.515$ as displayed in Table \ref{tab4}. This shows that we can obtain the channel-specific explanations using our proposed method without a loss in accuracy compared to other methods. 

\begin{table}[htb!]
\tabcolsep=0pt%
\TBL{\caption{Case Study 2: MJO  Classification Performance\label{tab4}}}
{\begin{fntable}
\begin{tabular*}{\textwidth}{@{\extracolsep{\fill}}lcccccc@{}}\toprule%

\TCH{Model} & \TCH{Accuracy} & \TCH{ +1/-1 Class Accuracy}\\\midrule

{\TCH{Standard Neural Network}}&  0.525 & 0.778\\
{\TCH{Prototype Network (No Channel-Specific Prototypes)}}&  0.515 & 0.792\\
\TCH{Prototype Network with Channel-specific Prototypes }& 0.553 & 0.805\\

\botrule
\end{tabular*}%
\end{fntable}}

\end{table}

\subsubsection{Local Explanations}
\begin{figure}[htb!]%
\FIG{\includegraphics[width=1\textwidth]{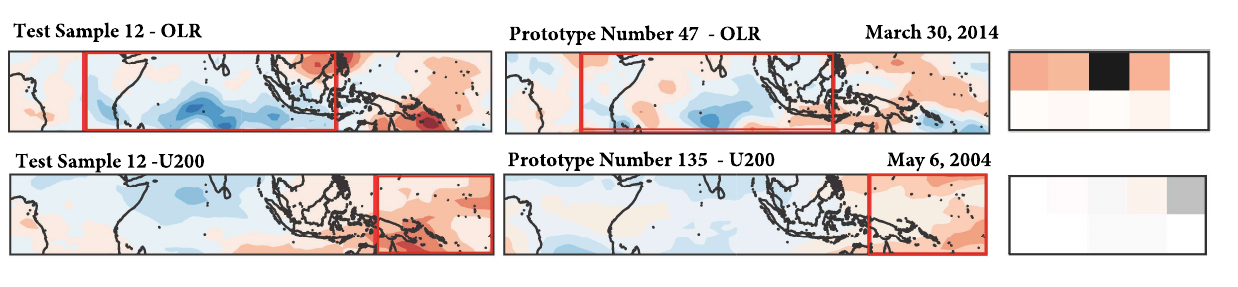}}
{\caption{Local explanation for Test Sample 12 (Label: MJO Phase 3). OLR (red=high OLR/suppressed convection, blue-low OLR/enhanced convection) and U200 (red=eastward direction, blue=westward direction) channels of test sample 12 are displayed on the left. Middle contains the highest scoring OLR prototype and second highest scoring U200 prototype that contributed to the model prediction within the red-outline boxes. Prototypes' associated location scaling grids on the right (grey=positive importance, red=negative importance) indicate the relevant locations within the original sample space of the respective prototype}
\textit{\textbf{Explanation of model prediction:} The model correctly predicts test sample 12 as Phase 3 primarily due to high similarity to negative OLR patterns over the Indian Ocean within the model identified prototype of Phase 3 from the March 30th 2014 training sample.}
\label{mjo_12}}
\end{figure}

We highlight examples of model predictions and associated local explanations within the test set, focusing on where the prototypes are associated with different channels and locations sourced from multiple training examples, a key difference in this method over standard prototype networks that only identify joint prototypes. For test sample 12 in Figure \ref{mjo_12}, the model correctly classified Phase 3 and identified a negative OLR pattern within the prototype (as indicated by the red-outline box in the top-middle) from the March 30th, 2014 training sample as the highest scoring prototype (score $2.213$) along with the associated location scaling grid that indicates the positive relevance (grey shading) of this pattern over the Indian Ocean region. In both the test sample and highest-scoring prototype (number 47), we can see a contrasting dipole of suppressed convection (as indicated by red shading where OLR is higher) over the maritime continent. These support the expected active convective patterns for MJO Phase 3 over the Indian Ocean \citep{wheeler2004all} indicating these explanations are in line with scientific reasoning. In summary, the model correctly predicts the test sample as Phase 3 due to high similarity to the OLR patterns identified in the previously seen March 30, 2014 MJO Phase 3 prototype.

The second highest contributing prototype (score $0.767$) captures a U200 zonal wind pattern flowing eastward over the Pacific Ocean as indicated by the location scaling grid away from the enhanced convection over the Indian Ocean. This is a key ability of the channel-specific prototype method to be able to source prototypes from different channels across different training samples and subsequently different locations. We also note that this prototype is sourced from a training image of the next phase 4. This may be due to the continuous nature of MJO phase activity where high scoring prototypes originate from adjacent phases. 

\begin{figure}[htb!]%
\FIG{\includegraphics[width=1\textwidth]{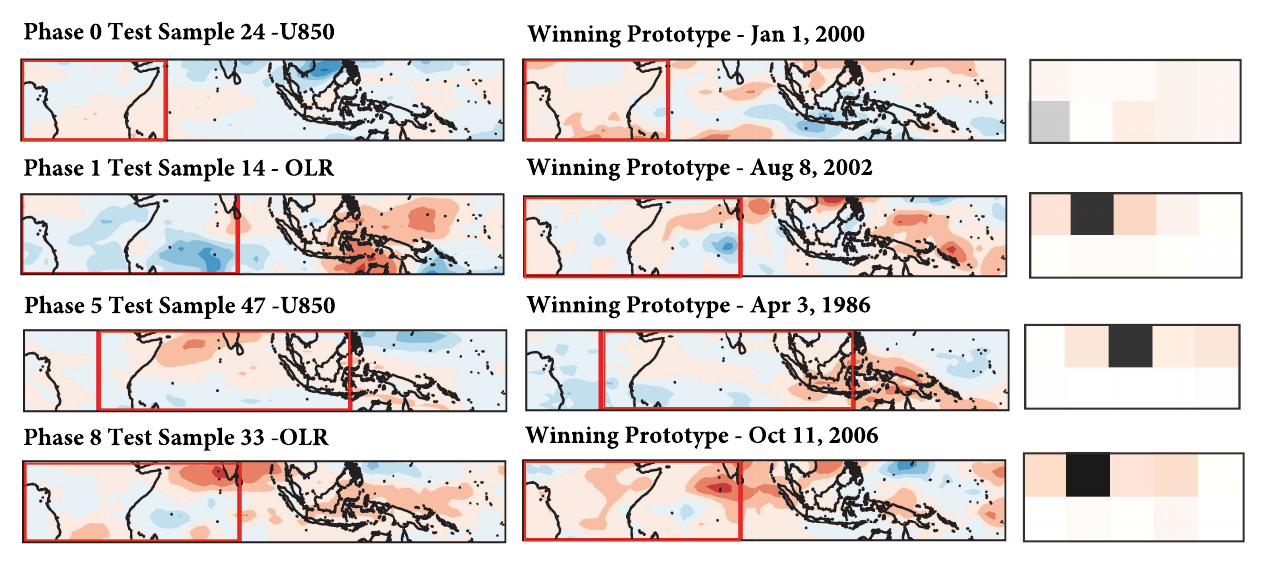}}
{\caption{Local explanation for test samples from Phases 0,1,5 and 8. One channel from each test sample is displayed in the left column. Corresponding channel prototypes with highest contributing scores are in the middle column displayed within the red-outline box (original training example is also displayed). OLR (red=high OLR/suppressed convection, blue-low OLR/enhanced convection) and U850 (red=eastward direction, blue=westward direction). Prototypes' associated location scaling grids on the right (grey=positive importance, red=negative importance) indicating the regional relevance of the prototype pattern within the test sample space.}
\label{fig:mjo_local_exps}}
\end{figure}

Furthermore, we display examples of local explanations produced by the model for multiple test samples from different MJO phases in Figure \ref{fig:mjo_local_exps}. In the top row of the image, the highest scoring prototype for Phase 0 or the `inactive' MJO phase does not capture any strong U850 patterns, only a weak eastward U850 zonal wind pattern over the African subcontinent. The location scaling grid does not assign a large importance to this pattern over this region as expected since the inactive phase does not correlate to any singular MJO pattern in a particular region. In the second row, the Phase 1 prototype captures the emergence of active MJO convection in the western Indian Ocean as also captured in the location scaling grid. In the third row, for Phase 5, zonal wind at 850mb flows eastward (indicated by red shading) towards the active convection over the Maritime continent as seen in the test sample and prototypical pattern captured. For the Phase 8 sample in the fourth row, the prototype captures the end of the MJO cycle where the suppressed convection is over the Indian Ocean. The identified prototypical pattern of suppressed convection shows the similarity to the positive OLR (red shading) over the western Indian Ocean in the test sample.

\subsubsection{Global Explanations}

\begin{figure}
    \centering
    \includegraphics[width=0.9\linewidth]{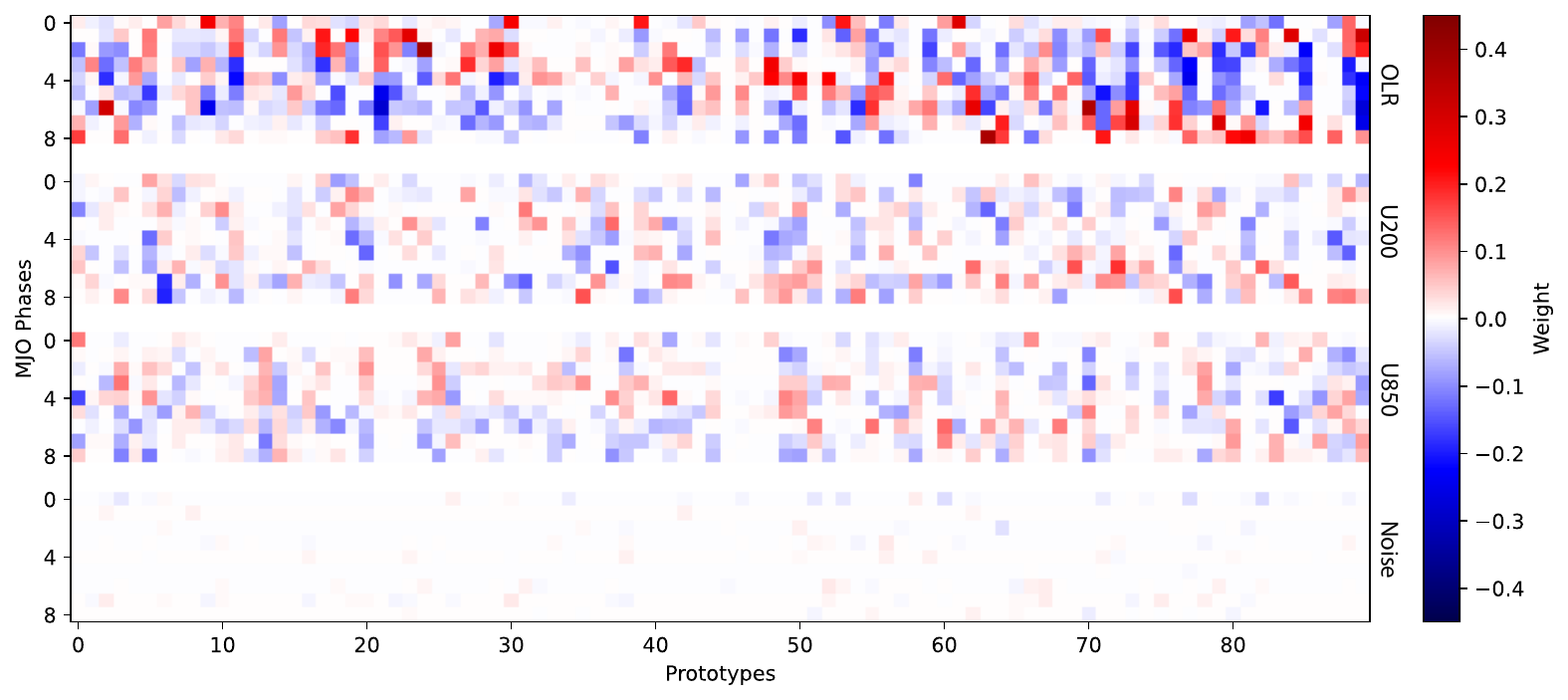}
    \caption{MJO case study (with added noise channel) final linear layer weight matrix (rows and columns permuted for visual clarity) with MJO phases 0-8 on the y-axis. The matrix shows the weights associated with the 90 prototypes from each channel (from top to bottom (separated by white dividers): OLR, U200, U850, Noise). Strong positive (red) and negative (blue) weights predominantly assigned to prototypes from the three environmental channels (OLR, U200, U850) while near zero (white) weights appear for prototypes from the Noise channel which is irrelevant to the MJO phase.}
    \label{fig:mjo_noise_weights}
\end{figure}

We further probe the relevance of individual channels on the dataset by adding a separate extra fourth channel to the MJO dataset. This fourth channel is filled with random noise sampled from a uniform distribution over $[0,1)$. We train a new identical model on this modified dataset and generate global explanations to explain model behavior when a random noise channel is added. In Figure \ref{fig:mjo_noise_weights}, the linear layer weight matrices for each channel, $\psi$, are displayed. we can see that the three channel prototype weights (OLR, U200, U850) show a large range of prototype weights. In contrast, the noise channel weights are near zero indicating they do not contribute highly to model predictions which is consistent with what we expect to see in the model behavior. This indicates the model is able to identify channels relevant to the overall task, consistent with results from the MNIST classification task.

\subsection{Case Study Discussion}

The channel-specific model for MJO phase classification successfully captures the variable dependent patterns that define MJO phases, performing comparably to both a standard neural network and a prototype network without channel separation. Class-wise model performance suggests the model is stronger at identifying region specific patterns for phase classification over the intensity of the pattern. Generated local explanations indicate the model learns the overall areas of convective activity associated with each phase (or an adjacent phase) and identifies regions of zonal flow towards and away from active convection areas. This reflects a key advantage of the channel-specific prototypes; allowing it combine convective and zonal wind pattern evidence drawn from distinct training examples and regions, providing a more richer explanation for model predictions. Ablation tests with additional random noise channels indicate the model can identify relevant channel prototypes and limit the use of non-informative channel prototypes by weighting the linear layer parameters accordingly. Overall, these results showcase that channel-specific prototypes are well suited for tasks where spatial patterns vary across variables and regions.

\section{Remote Sensing Land-Use Classification Case Study}\label{sec:EO}
\subsection{Dataset and Task}
We demonstrate our proposed method on the Euro-SAT satellite ML dataset \citep{helber2019eurosat} which is a common multi-spectral remote sensing benchmark dataset. The Euro-SAT dataset consists of 28,000 satellite snapshot images captured by Sentinel-2 across the European region with 10 land-use classification labels: industrial, residential, annual crop, permanent crop, river, sea and lake, herbaceous vegetation, highway, pasture, and forest. Sentinel-2 captures images across 13 spectral bands, each with relevant information that can indicate specific characteristics of the Earth's surface (vegetation cover, urban settings, water bodies, etc.). The spectral bands captured are: Aerosols, Blue, Green, Red, Red Edge 1, Red Edge 2, Red Edge 3, Near Infrared (NIR), Red Edge 4, Water Vapor, Cirrus, Short Wave Infrared (SWIR) - 1, and  Short Wave Infrared (SWIR) - 2. 

\subsection{Model Architecture and Training}
We use the standard 60-20-20 train-validation-test split provided for the dataset in torchgeo Python package \citep{stewart2024torchgeo} from \citet{neumann2019domain} for the EuroSAT Land Classification Dataset. Each land patch image is of size $64 \times 64$ pixels with 13 channels. Each channel is separated and fed individually into the encoder which consists of four sequences of convolution, leaky ReLU activation and average pooling layers. For each of the ten available classes, we find 4 prototypes per channel. Across all classes, this results in 40 prototypes per channel and 520 prototypes in total. In this classification task, the absolute location of spatial patterns within an image are not directly relevant to the label, so we omit learning the location scaling parameter to allow prototypes to occur in any region within the image for location-agnostic predictions (and explanations). The stage two prototype projection step is completed every 4 epochs or 2 cycles. Column 3 in Table \ref{tab2} outlines the model training parameters. Model hyperparameters were minimally tuned manually.

\subsection{Results}
\subsubsection{Model Performance}
For the EuroSAT land-use classification task, the channel-specific prototype model achieved a performance accuracy of $0.926$. The prototype network where prototypes are learnt jointly achieves an accuracy of $0.932$. The class-wise accuracy ranged from $0.820-1.0$ accuracy. A standard black box CNN achieves $0.945$ performance accuracy similar to the accuracy stated in \citet{helber2019eurosat} for the EuroSAT land-use classification task. Model performance indicates the channel-specific prototype method achieves similar performance across model comparisons. 

Above, we have investigated generating channel-specific prototypes trained alongside the model encoder, $E$, in which prototypes are simultaneously learnt with the encoder weights, $\theta_E$ within the same training procedure. In addition, we evaluate whether prototypes-based explanations can be generated from a separately trained black box encoder via transfer learning from pre-trained models. We perform this experiment to examine cases where users may already have a pre-trained network for their task for which they would like to use prototypes to generate explanations. This is similar to how one would approach using a post-hoc XAI method for a pre-trained model though different in that additional training is necessary to learn the appropriate prototypes. We use the same 4-layer CNN architecture mentioned above as the encoder with a standard linear classifier and train it on the land-use classification task, similar to standard neural network model and without the use of any prototypes. We then transfer the learnt pre-trained encoder weights and apply the prototype layer on this pre-generated embedding to learn prototypes in two different encoder weight settings \textit{frozen encoder} and \textit{unfrozen encoder} rather than learning both the encoder and prototypes simultaneously. 

In the first experiment, we freeze the encoder weights and learn the prototype parameters with the fixed embedding (i.e.\ in Stage 1, $\theta_E$ is initialized with the pre-trained embedding and frozen while the prototypes, $\phi$, are learnt). In the second experiment, we allow the encoder weights to be fine-tuned while training the prototype parameters (i.e.\ in Stage 1, $\theta_E$ is initialized to the pre-trained weights and both $\theta_E$ and $\phi$ parameters are learnt). The results of both experiments are shown in the last row of Table \ref{tab3}. In both cases (frozen and unfrozen encoder weights) the model learns the appropriate prototypes with a pre-trained embedding with similar accuracy to prototype network with channel-specific prototypes that was trained from scratch. The unfrozen setting results in a slightly higher performance over the frozen setting since the encoder is further fine-tuned in the training process, resulting in a more accurate embedding for the prototypes. This indicates that using a separately trained encoder on a task and subsequently applying prototypes could result in similar performance to learning them simultaneously, which implies one can use the intrinsically interpretable prototype model without needing to train the entire model (encoder) from scratch.

\begin{table}[htb!]
\tabcolsep=0pt%
\TBL{\caption{Case Study 3: Land Use Classification Performance\label{tab_land}}}
{\begin{fntable}
\begin{tabular*}{\textwidth}{@{\extracolsep{\fill}}lcccccc@{}}\toprule%

\TCH{Model} & \TCH{Performance Accuracy} \\\midrule

{\TCH{Standard Neural Network}}&  0.945\\
{\TCH{Prototype Network (No Channel-Specific Prototypes)}}&  0.932\\
\TCH{Prototype Network with Channel-specific Prototypes }& 0.926\\
\TCH{Pre-trained Prototype Network with Channel-specific Prototypes }& 0.870 (frozen),  0.917 (unfrozen) \\
\botrule
\end{tabular*}%
\end{fntable}}

\end{table}

\subsubsection{Local Explanations}

\begin{figure}[htb!]%
\FIG{\includegraphics[width=1\textwidth]{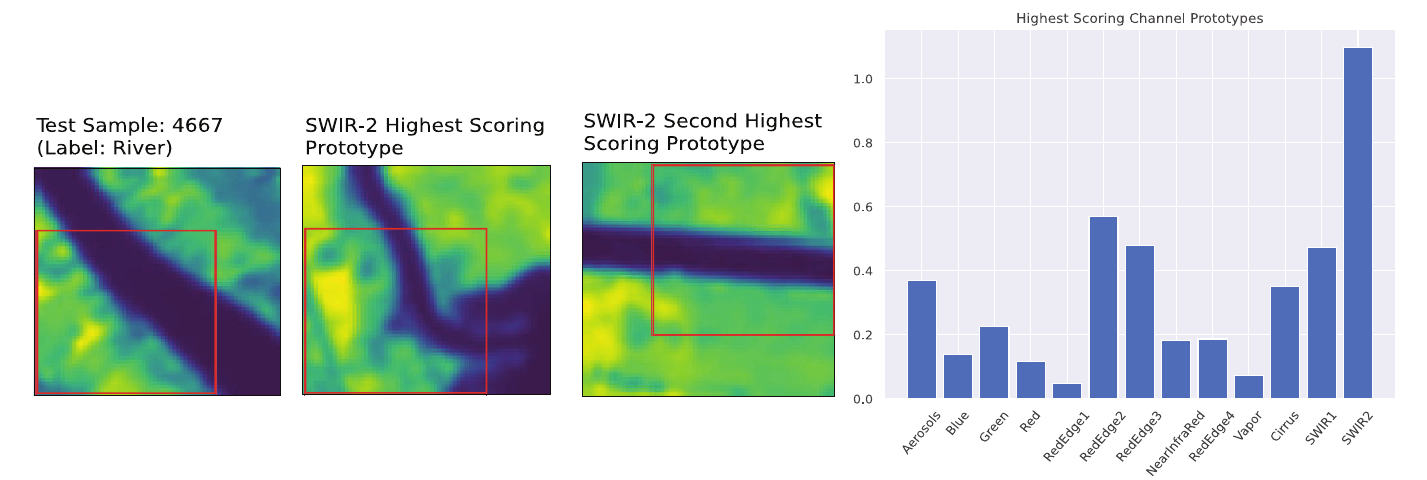}}
{\caption{Local prototype-based explanation Test Sample 4667 (Label: River). The Short Wave Infrared (SWIR)-2 channel of the test sample is shown on the left. The highest scoring prototype from the SWIR-2 channel is shown in the second panel (red outline) with a similarity score of 1.099; for visual reference, the corresponding training image from which the prototype is derived is also displayed. The second highest scoring SWIR-2 prototype is shown third from the left with a score of 0.576. The rightmost panel summarizes the highest prototype similarity scores from each channel for this instance prediction. \\ \textit{\textbf{Explanation of model prediction:} The model predicts that the image belongs to class `River' because of a high similarity to the SWIR-2 channel of a prototype with the class label `River', which makes the largest contribution to the final prediction.}}
\label{fig:rs_4667}}
\end{figure}

We examine the generated evidence for correctly classified test sample number 4667 with the label; River in Figure \ref{fig:rs_4667}. Channel 12 (SWIR-2) of test sample 4667 is shown in the leftmost part of Figure \ref{fig:rs_4667}. Adjacently to the right, the figure shows the highest scoring prototype for this instance's classification with a score of $1.009$. This particular prototype originates from channel 12 of a training image sharing the same class label, River. In Figure \ref{fig:rs_4667}, we show the second highest scoring prototype with a contributing score of $0.576$, also from SWIR-2 channel of a training image sharing the same class label, River. Visually, one can see the similarity between the pattern identified in the test sample as well as in the prototypical patterns identified in the high scoring prototypes. A breakdown of each channel's highest scoring prototype's contribution to the final prediction in Figure \ref{fig:rs_4667} show that the highest scoring prototype from channel 12 (SWIR-2) contributed the most to this prediction. This is in line with what we expect to see as water bodies may be more visible in short-wave infrared channels \citep{huang2018detecting}. Both RedEdge2 and RedEdge3 channels provided prototypes with similar scores to the second highest scoring SWIR-2 prototype. In comparison, other channels' highest scoring prototypes were minimal in their contributions. For more examples of model generated prototypes across the ten land-use classes see Figure S3 in Supplement Section 4.

\subsubsection{Global Explanations}
\begin{figure}[htb!]
\centering
\begin{subfigure}{.5\textwidth}
  \centering
  \includegraphics[width=.9\linewidth]{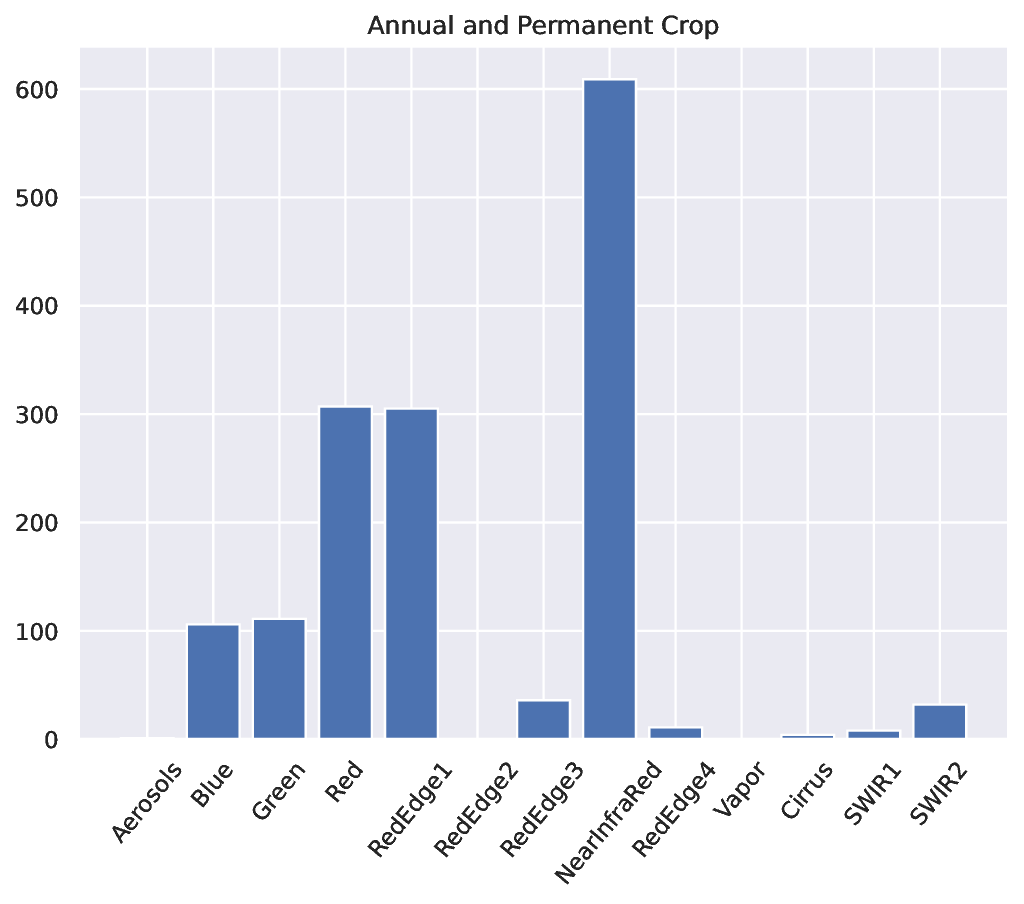}
  \caption{Annual and Permanent crop}
  \label{fig:rs_crop}
\end{subfigure}%
\begin{subfigure}{.49\textwidth}
  \centering
  \includegraphics[width=.9\linewidth]{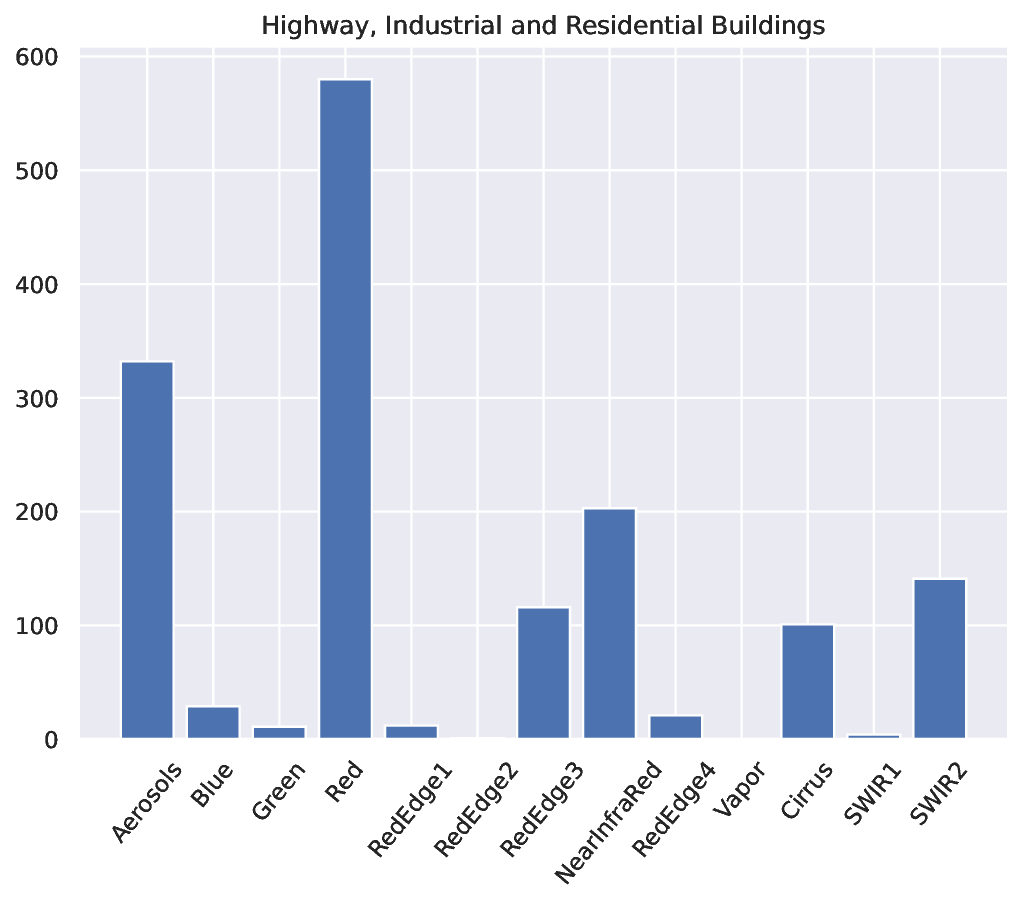}
  \caption{Highway, Industrial and Residential buildings}
  \label{fig:rs_building}
\end{subfigure}

 \medskip
\begin{subfigure}{.5\textwidth}
  \centering
  \includegraphics[width=.9\linewidth]{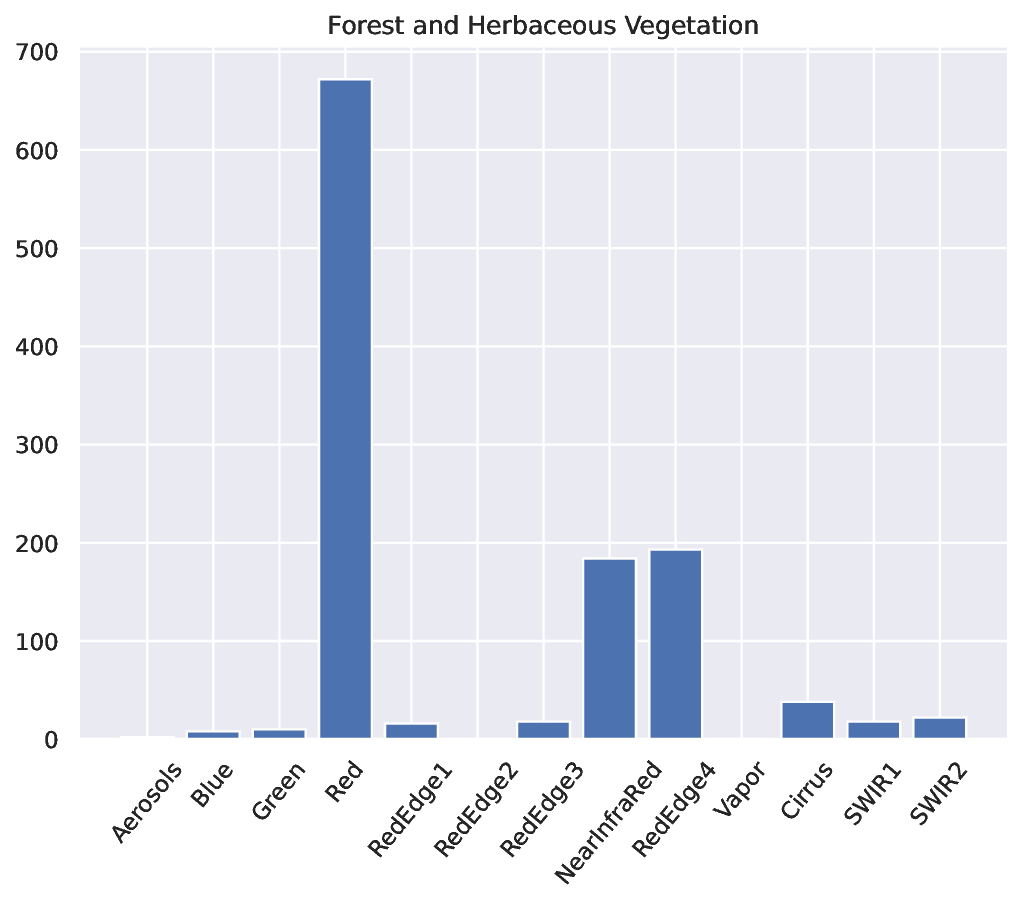}
  \caption{Forest and Herbaceous vegetation}
  \label{fig:rs_forestveg}
\end{subfigure}
\begin{subfigure}{.49\textwidth}
  \centering
  \includegraphics[width=.9\linewidth]{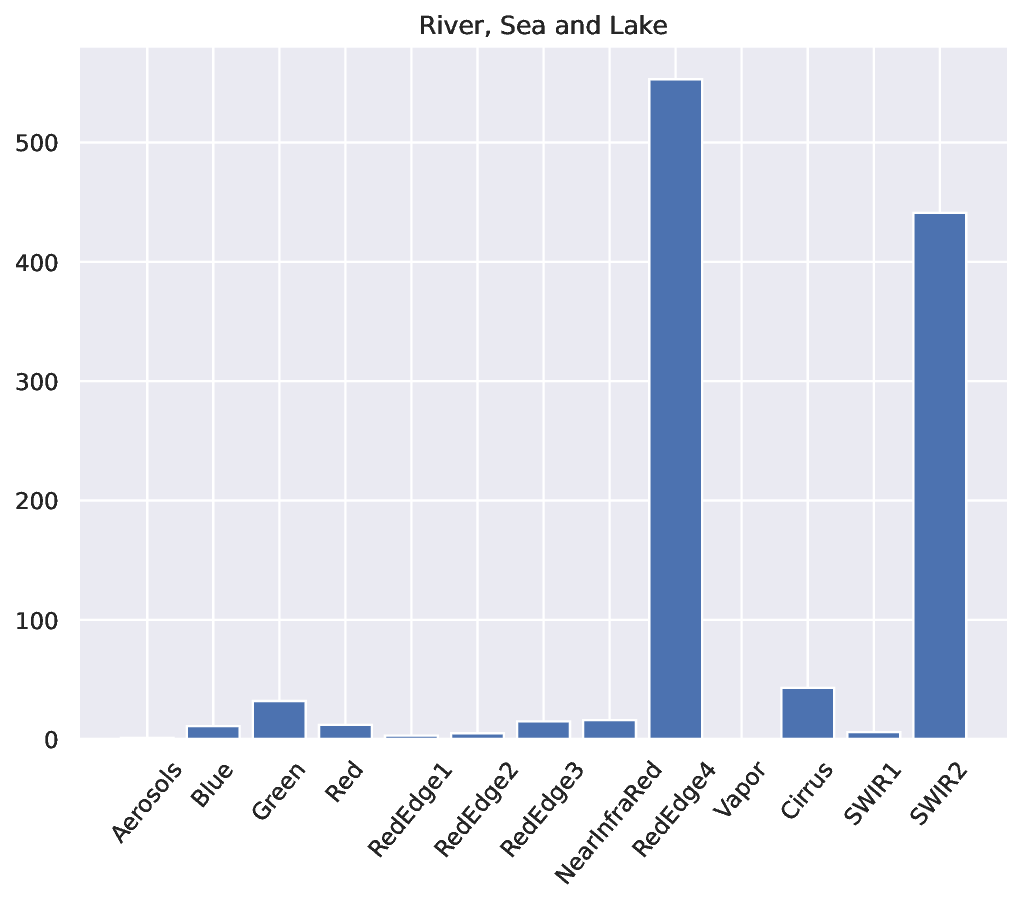}
  \caption{River, Lake and Sea}
  \label{fig:rs_riversea}
\end{subfigure}
\caption{Across the test set, the frequency with which each spectral channel provides the top-scoring prototype in correctly classified instances from (upper left) Annual and Permanent Crop classes, (upper right) Highway, Industrial, and Residential Building classes, (lower left) Forest and Herbaceous Vegetation classes, and (lower right) River, Sea, and Lake classes.}
\label{fig:rs_global}
\end{figure}

For a remote sensing task with a large number of spectral channels, it may be useful to investigate the importance of specific channels for band selection purposes. By identifying the most frequent highest scoring prototypes, we can analyze the specific channels that may be more important in this learning task, thus gaining an understanding of the overall model reasoning. In Figure \ref{fig:rs_global}, we conduct an analysis for different groups of class labels and find that specific channels contribute more high scoring prototypes for specific classes over others, corresponding to their relevance for the class. For example, for the agriculture related class labels such as annual crop, permanent crop and pasture, we can see the channels from which the highest scoring prototypes came from the Infrared, Red, and RedEdge channels. Fewer high scoring prototypes came from the Aerosols, Vapor and Cirrus channels in Figure \ref{fig:rs_crop}. Typically the red, infrared and red edge channels are used to calculate vegetation indices to estimate crop yield \citep{kang2021crop, al2025assessment} and chlorophyll content \citep{chang2010estimating}. In Figure \ref{fig:rs_building}, for manmade structure-related classes such as highways and industrial and residential buildings, high scoring prototypes came from the red and aerosols channels. Most high scoring prototypes for the forest and herbaceous vegetation classes come from the red or infrared channels as seen in Figure \ref{fig:rs_forestveg}. For water-related classes such as river and sea and lake, most high scoring prototypes come from the SWIR-2 and infrared channels. As discussed earlier, SWIR bands can capture water-based properties better than other spectral channels which is showcased in Figure \ref{fig:rs_riversea} \citep{huang2018detecting}. For the same channel importance analysis conducted for the entire test set across all class labels, see Figure S4 in Supplement Section 5. These results clearly indicate that specific classes are clearly dependent on a select few channels within the dataset, dependent on how the surface properties of classes are captured by different spectral bands.

\subsection{Case Study Discussion}
Results show the model identifies prototypical patterns from a select few highly relevant channels to make predictions with minimal drops in performance compared to both standard neural networks and prototype networks without channel-specific prototypes. At the instance-level, the model consistently chooses prototypes from specific channels whose spectral characteristics most clearly capture class-relevant land-use features. Globally, we see this trend for groups of similar class labels, where specific channels are more frequently selected (e.g SWIR-2 bands that are sensitive to water-related land-use classes, Red and Infrared bands for forest and vegetation based classes) that is in line with expected channel properties. We also find that prototypes can easily be generated from a pre-trained encoder with similar performance levels to training the prototype networks from scratch, increasing computational efficiency.

\section{Discussion} \label{Discussion}

In this study, we develop an intrinsically interpretable prototype-based neural network method that extracts channel-specific patterns from geoscientific datasets as the basis for explanations. We showcase this method on synthetic case study as well as two geoscientific use cases: an MJO phase classification task and a remote sensing land-use classification task. In each task, we train a model and generate local, instance-based explanations and global, model-level explanations to understand model reasoning with comparable performance accuracy to standard neural networks.

Across the three case studies, our results consistently demonstrate that channel-specific prototypes provide an effective and flexible mechanism for interpretability in deep learning models trained on multi-channel datasets. In the synthetic MNIST digit classification task, we designed a channel-dependent labeling scheme to demonstrate that our proposed method correctly identifies prototypical matching digit patterns within the relevant channels, while assigning little weight to channels that, by construction,  contain no class-relevant information. We then applied our proposed method to more complex, real-world scientific datasets to illustrate the added scientific value that channel-specific prototypes can provide. 

In both real-world case studies, channel-specific prototypes yield physically meaningful local explanations. In the MJO phase classification task, instance-level explanations are particularly valuable through the method's ability to identify prototypes from distinct training examples and spatial regions that exhibit convective and dynamic spatial patterns within individual variables. This leads to more physically meaningful explanations than standard prototype networks, which can not isolate patterns tied to a single physical variable. Additionally, prototypes from adjacent MJO phase labels often contribute high similarity scores, reflecting the continuous nature of the MJO phase cycle. In the remote sensing task, instance-level explanations reveal that certain channels contribute higher prototypes scores for specific classes (e.g SWIR-2 for River), indicating the model selectively relies on channels that better capture land-surface characteristics for a specific prediction instance. Together, these results show that channel-specific prototypes support comprehensive local explanations by identifying relevant patterns within specific channels that directly drive model predictions.

While most prototype-based XAI architectures focus on generating local explanations for specific instances, a key advantage of our proposed channel-specific model is the ability to generate global explanations through aggregate analysis of channel-wise prototypes across the dataset. By examining the final layer weight matrix, we can infer which channels are generally most relevant to the learning task (i.e the channels assigned higher weights), supporting potential feature or band selection. In the MNIST example, the model assigned near-zero weights to prototypes from the channel containing irrelevant patterns. In contrast, the MJO task exhibits relatively uniform prototype weights across the environmental variable channels, indicating all three contribute meaningfully to phase classification. For the Euro-SAT land-use task, we observe substantial variation in the channels that provide high scoring prototypes, identifying specific sets of relevant channels for different class labels across the test set. These global explanations highlight how aggregate prototype usage reveals consistent and interpretable relationships between specific spectral bands and land-cover classes, particularly valuable for multispectral datasets where different band combinations are critical for detecting specific land surface features. 

As prototypes are learned as part of model optimization, predictions are explicitly made from similarity values between the test sample and learnt prototypes, yielding explanations that directly reflect the model's internal decision process. In contrast to post hoc XAI methods, the explanations produced by our method do not depend on externally applied measures and therefore do not produce multiple varied explanations for the same model prediction \citep{mamalakis2022investigating}. While different prototype configurations may result in different representative prototype examples, the resulting explanations remain faithful and directly tied to the model's learned reasoning at both instance and global levels.

Similar to other prototype-based networks, our channel-specific method comes with additional training complexity compared to standard neural networks. Beyond tuning conventional model hyperparameters, manual adjustment is often required to generate meaningful prototypical examples, including adjusting the number of prototypes needed per channel, the frequency of prototype projection, the weighting of the loss terms as well as the strength of the $L1$ regularization. In particular, balancing loss-term coefficients strongly affects both model convergence and model performance. In practice, the cluster cost is typically weighted equal to or higher than the separation cost, allowing the model to prioritize learning representative prototypes for a class. The diversity loss term is usually assigned a lower weight to prevent the prototype redundancy. Setting the number of prototypes per channel is primarily task-dependent and may rely on domain-specific knowledge to tune. Setting too few prototypes usually results in poor model performance. Increasing the number of prototypes generally allows greater representational flexibility, though too many prototypes per channel can hinder the model from identifying representative samples, increase the training time during the projection step, and result in repetitive prototype explanations. Within a reasonable range, the frequency of prototype projection has little impact on model accuracy, though it should remain relatively low (every 2-3 training cycles) to allow the model to converge on a set of meaningful prototypes for the learning task.

Scalability also presents potential challenges for prototype-based methods. Unlike post hoc XAI methods whose computational demands are largely independent of the number of features/channels in the training dataset, the number of prototypes in channel-specific models grows linearly with the increase in number of channels. This can introduce both interpretability and computational challenges, which can be a training bottleneck for larger datasets especially during the projection step onto the entire training dataset. Increasing the $L1$ regularization as well as adjusting the number of prototypes per channel can help constrain the model to focus on fewer, more informative prototypes. Future work could explore projecting prototypes onto representative sub-samples of the training dataset or adopting other alternative prototype selection strategies to reduce memory or runtime demands. 

Finally, we demonstrate that channel-specific prototypes can be efficiently learned using pre-trained encoders, as shown in the remote sensing case study. Leveraging a frozen pre-trained encoder allows the model to retain previously learnt feature representations while learning prototypes for interpretability. Freezing the pre-trained encoder led to only an approximately \~4\% reduction performance in comparison to the jointly fine-tuning both the pre-trained encoder weights and prototypes, which is expected, given the reduced optimization flexibility. Our approach offers a workflow similar to standard post hoc XAI methods while retaining the advantage of intrinsic interpretability prototype-based reasoning. Moreover, this approach can reduce computational costs by enabling explanations without retraining the full model from scratch. 

In conclusion, our proposed channel-specific prototype-based XAI method enables deep learning models to identify interpretable channel-specific prototypical patterns as necessitated by geoscientific raster datasets that distribute relevant information across multiple channels. Across all case studies, our method attains comparable performance scores to standard deep learning models and standard prototype networks without channel-specific explanations while additionally supporting both local-level and global-level explanations. By allowing the model to learn prototypical patterns within individual channels, it provides more detailed insights into which channels drive specific predictions at the instance level and channel-class relationships at the global level. More broadly, advancing intrinsically interpretable deep learning methods to provide richer explanations for geoscientific tasks supports the development of trustworthy, transparent and accessible deep learning models for geoscientific research and decision-making. 

\begin{Backmatter}

\paragraph{Acknowledgments}
This research was conducted using computational resources and services at the Center for Computation and Visualization, Brown University.

\paragraph{Funding Statement}
This work was partially supported by the SciAI Center, funded by the Office of Naval Research (ONR), under Grant Number N00014-23-1-2729. We are grateful to an anonymous donor for their contribution to support the work of early-career researchers impacted by award terminations. 

\paragraph{Competing Interests}
The authors declare none.

\paragraph{Data Availability Statement}
The data that support the findings of this study is freely available. Raw data used in the MJO classification task can be found at \url{https://doi.org/10.5281/zenodo.3968896} \citep{toms2020testing_code}. Raw data used in the land use classification task is obtained from TorchGeo package \citep{stewart2024torchgeo} at \url{https://torchgeo.readthedocs.io/en/latest/api/datasets.html#torchgeo.datasets.EuroSAT}. Processed data and saved models used in this study can be found at \url{https://doi.org/10.5281/zenodo.18425035} \citep{narayanan_data}. Code used to generate synthetic MNIST data and study results are located in a GitHub repository \url{https://github.com/anaray23/ChannelSpecific_PrototypeNetwork}.

\paragraph{Ethical Standards}
The research meets all ethical guidelines, including adherence to the legal requirements of the study country.

\paragraph{Author Contributions}
Conceptualization: AN, KB; Data Curation: AN; Methodology: AN, KB;  Software: AN; Visualization: AN;  Writing - Original draft: AN, KB. Supervision and Funding acquisition: KB. All authors approved the final submitted draft.


\printbibliography

\end{Backmatter}

\end{document}


\begin{Frontmatter}

\title[Article Title]{Supplement}

\end{Frontmatter}

\section{Model Comparison with Separate Encoder Architecture} \label{supp:sep_encoders}

Our study uses a shared encoder architecture set-up where one singular encoder function, $E$, is applied to each channel of the input data separately. For comparison, we evaluate an architecture variant where each channel is processed by its own separate encoder function, $E_j$, each with separate trainable parameters, $\theta_j$, increasing the total number of encoder trainable parameters by a factor equal to the number of channels. We find that despite increasing the number of encoder trainable parameters, model performance across all three case studies did not improve compared to the shared-encoder set-up, motivating our use of the shared-encoder architecture throughout the study. 

\begin{table}[htb!]
\tabcolsep=0pt%
\TBL{\caption{Performance Comparison of Shared Encoder vs. Separate Encoder Model Architecture. Case Study 1 refers to the Synthetic MNIST case study. Case Study 2 refers to the Madden-Julian Oscillation case study. Case Study 3 refers to the Land Cover Classification case study.}}
{\begin{fntable}
\begin{tabular*}{\textwidth}{@{\extracolsep{\fill}}lcccccc@{}}\toprule%

\TCH{Model} & \TCH{Case Study 1} & \TCH{Case Study 2} & \TCH{Case Study 3}\\\midrule

{\TCH{Shared Encoder Model (used in study)}}&  0.978 & 0.553 & 0.926\\
{\TCH{Separate Encoder Model}}&  0.968 & 0.543 & 0.916\\

\botrule
\end{tabular*}%
\end{fntable}}
\end{table}

\section{Local Explanation for MNIST Sample 98 using Prototype Network without Channel-Specific Prototypes} \label{supp:mnist_98_joint}

\begin{figure}[htb!]
\centering
\begin{subfigure}{.49\textwidth}
  \centering
  \includegraphics[width=.8\linewidth]{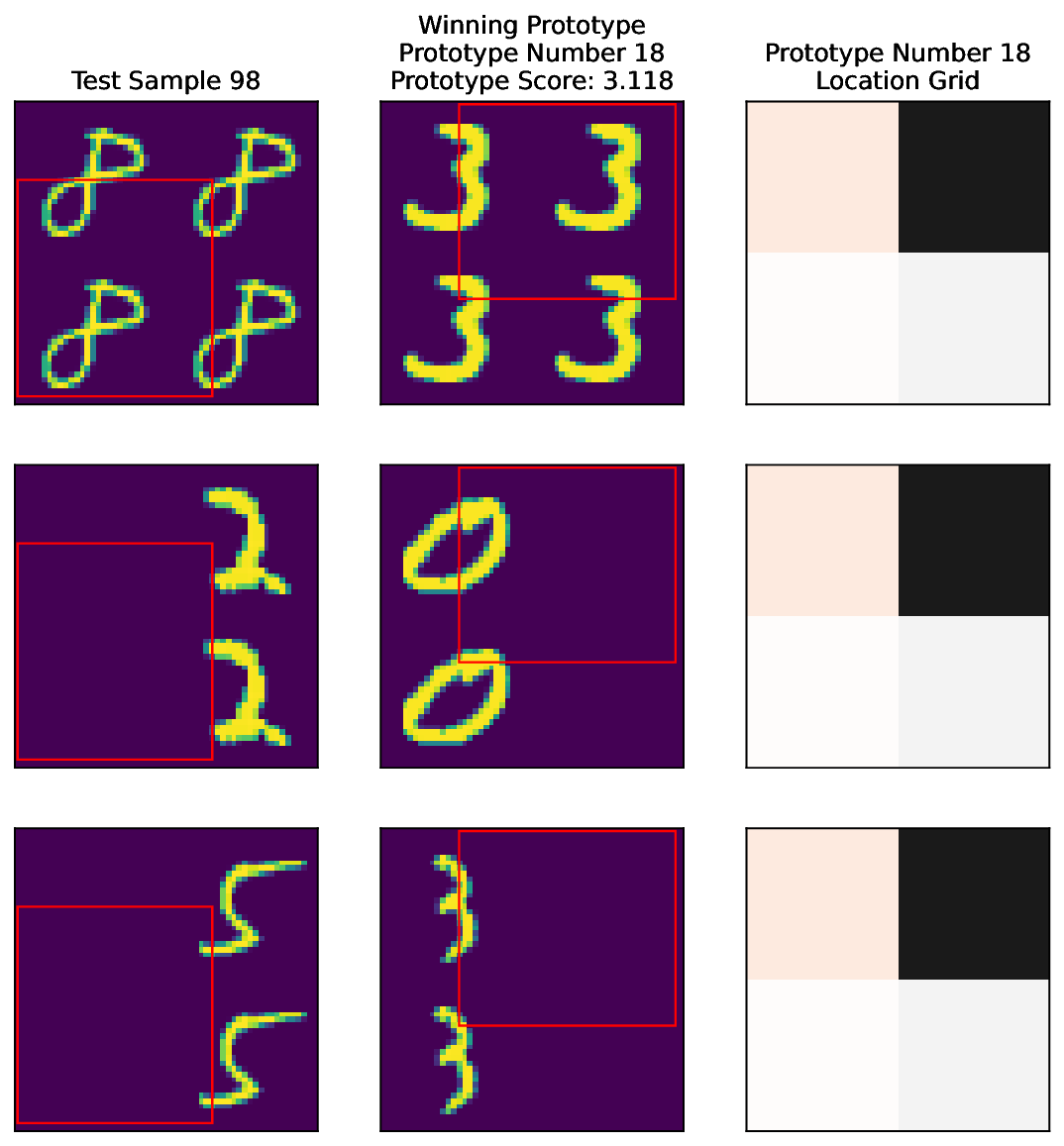}
  \caption{\centering Prototype Network \\(Predicted Label: 3, True Label: 2)}
  \label{supp_fig:mnist_98nc}
\end{subfigure}
\caption{Local Explanation for Test Sample 98 with each row corresponding to three image channels. Test sample is shown in the left in both sub-figures with red-outline box indicating the part of the image that had highest similarity to the prototype. Highest scoring prototypes are shown in the middle column with the prototype receptive field indicated in the red outlined box and associated location scaling in the right column (red indicates negative relevance, grey indicates positive relevance).}
\end{figure}

We examine the prediction for MNIST test sample 98 using a prototype network without channel-specific prototypes in Figure \ref{supp_fig:mnist_98nc}. For this test sample, the network incorrectly predicts class ``3''. Unlike the channel-specific approach, the learned prototype aggregates patterns jointly across all three channels from a single training example rather than isolating channel-wise patterns from multiple trainin examples. The highest scoring prototype, originating from training sample 4659, contains the digit patterns ``3'', ``0'' and ``3'' across the three channels respectively, however the model primarily identifies the `blank' regions as the salient prototypes. Specifically, blank patterns on the right hand side of the prototype associated with class ``3''  drive the incorrect prediction. The model identifies similarity between these blank regions in the prototype and corresponding blank regions in the test sample as the primary driver of the prediction. This reliance likely arises because the model was not able find a class ``2” prototype that simultaneously matches all digit patterns present in the test sample across channels. Consequently, because prototypes span all channels jointly, the model fails to capture the relevant pattern in channel 2 while disregarding irrelevant channels. This underscores the increased flexibility of the channel-specific method, particularly when some channels contain little to no class-relevant information.

\section{MJO Case Study: Accuracy Confusion Matrix}\label{supp:MJO_cm}

\begin{figure}[htb!]%
\FIG{\includegraphics[width=0.5\textwidth]{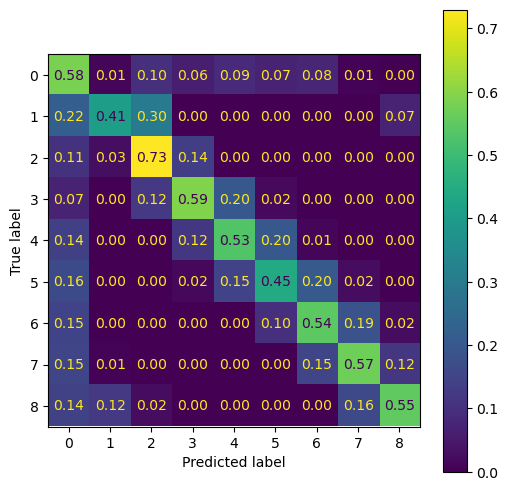}}
{\caption{Confusion Matrix of Individual Class Accuracy with Channel-Specific Prototypes for MJO Classification Task}
\label{supp_fig:mjo_cm}}
\end{figure}

\section{Examples of Model Identified Prototypes for Remote Sensing Land-Use Classes} \label{supp:rs_prototype_examples}

For the Euro-SAT land classification dataset, we showcase examples of top three most frequently high scoring prototypes for each of the 10 land-use classes within the red-outlined boxes. The original training image from which the prototypes originate are also shown.

\begin{figure}[htb!]%
\FIG{\includegraphics[width=0.8\textwidth]{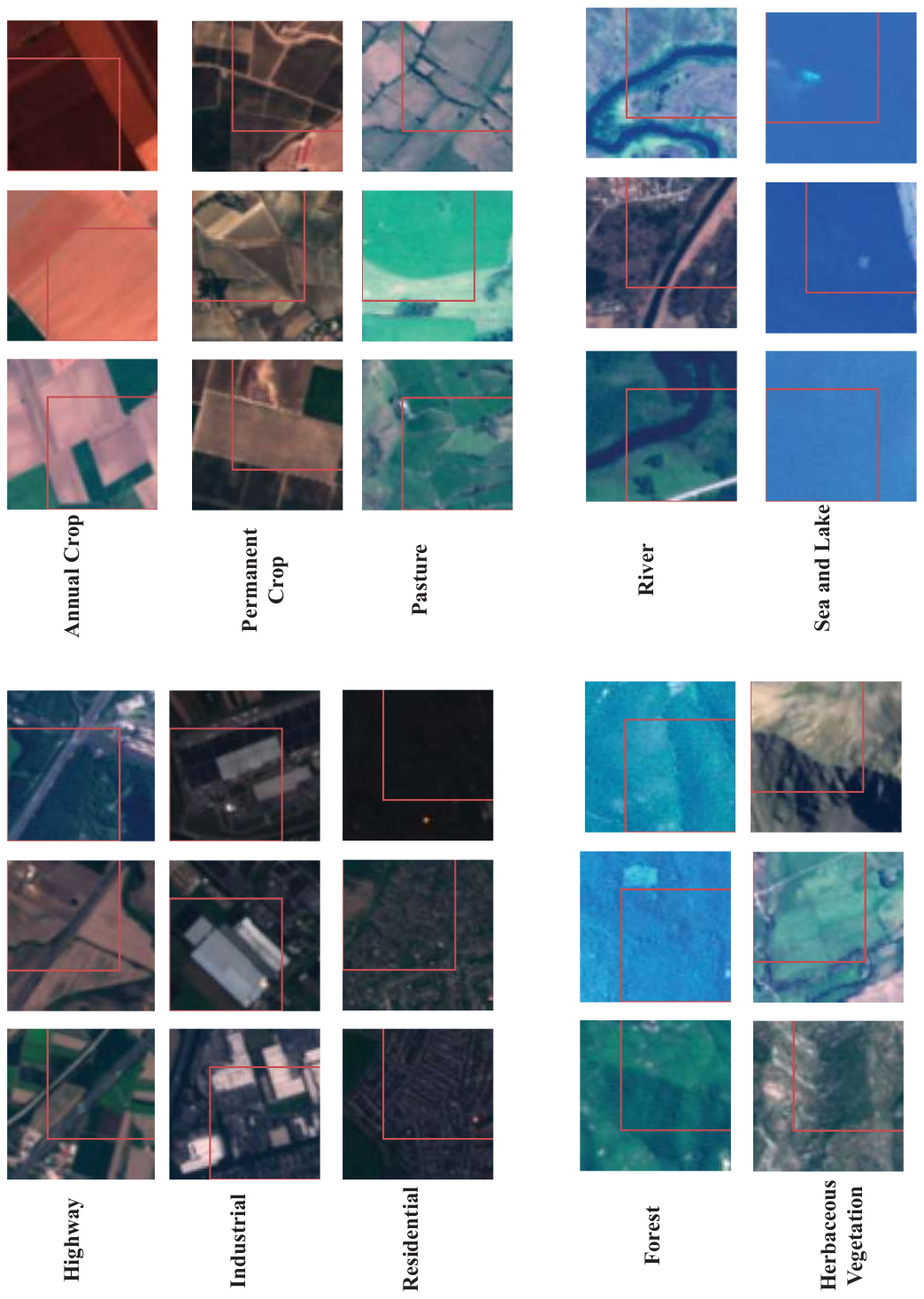}}
{\caption{Examples of the top three frequent high scoring prototypes for each of the EuroSAT land-use classes (RGB channels shown)}
\label{supp_fig:rs_proto_examples}}
\end{figure}

\section{Channel-wise Number of Highest Scoring Prototypes Across Remote Sensing Test Set} \label{supp:rs_protos_testset}
\begin{figure}[htb!]%
\FIG{\includegraphics[width=0.5\textwidth]{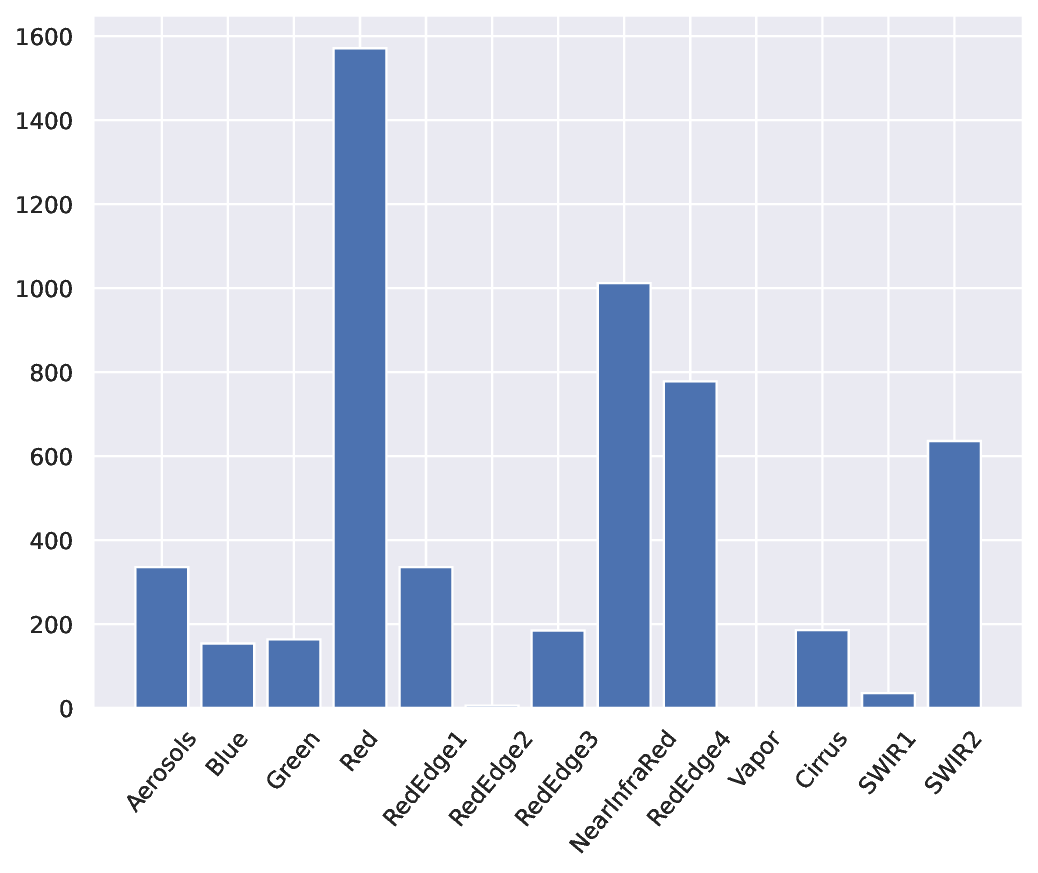}}
{\caption{Number of times each channel provided the highest scoring prototype for the correctly classified samples in the test set}
\label{supp_fig:rs_all}}
\end{figure}

Across the test set, most of the highest scoring prototypes come from the RGB channels as well as the Infrared, Red Edge4 and SWIR-2 channels as seen in Figure \ref{supp_fig:rs_all}. Very few, if not zero, high scoring prototypes came from the aerosol, red edge 1-3 or vapor channels. This could likely arise because the aerosol channel which denotes aerosol levels in the atmosphere at the time of the image capture should not have any effect on the land-use class within the image. Similar reasoning could be made for the vapor channel.

\begin{backmatter}
\printbibliography
\end{backmatter}